\documentclass{article}

\usepackage{microtype}
\usepackage{graphicx}
\usepackage{booktabs} 
\usepackage{xcolor}

\usepackage{hyperref}
\usepackage{url}

\usepackage[accepted]{mlsys2025}

\usepackage{amsmath,amsfonts,bm}

\def\eqref#1{equation~\ref{#1}}

\def\1{\bm{1}}

\DeclareMathAlphabet{\mathsfit}{\encodingdefault}{\sfdefault}{m}{sl}
\SetMathAlphabet{\mathsfit}{bold}{\encodingdefault}{\sfdefault}{bx}{n}

\usepackage{amsmath,amssymb,amsthm}
\usepackage{multirow}
\usepackage{array}
\usepackage{enumitem}
\usepackage[ruled,linesnumbered,algo2e]{algorithm2e}
\usepackage{fvextra}   

\newif\ifdraftmode
\draftmodetrue          
 
\definecolor{phred}{RGB}{190,30,45}
\definecolor{prelblue}{RGB}{20,90,170}
\definecolor{todogreen}{RGB}{0,120,60}
 
\ifdraftmode

  \newcommand{\PH}[1]{\textcolor{phred}{\textbf{#1}}}
  
  \newcommand{\todo}[1]{\textcolor{todogreen}{\small[TODO: #1]}}
  \newcommand{\draftnote}[1]{\textcolor{todogreen}{\small[#1]}}
\else
  \newcommand{\PH}[1]{\errmessage{PLACEHOLDER SURVIVED INTO FINAL BUILD}}
  
  \newcommand{\todo}[1]{}
  \newcommand{\draftnote}[1]{}
\fi
 
\newcommand{\Dfit}{\mathcal{D}_{\mathrm{fit}}}      
\newcommand{\Dcid}{\mathcal{D}_{\text{cal-ID}}}     
\newcommand{\Dcext}{\mathcal{D}_{\text{cal-EXT}}}   
\newcommand{\Dtest}{\mathcal{D}_{\mathrm{test}}}    
\newcommand{\Dood}{\mathcal{D}_{\mathrm{OOD}}}      

\mlsystitlerunning{Structural Uncertainty for LLM-Guided Symbolic Regression under Extrapolation}

\begin{document}

\twocolumn[
\mlsystitle{Know When the Equation Is Wrong: Structural Uncertainty for LLM-Guided Symbolic Regression under Extrapolation}

\begin{mlsysauthorlist}
\mlsysauthor{Mostafa Cham}{umbc}
\mlsysauthor{Bayu Adhi Tama}{umbc}
\mlsysauthor{Jianwu Wang}{umbc}

\end{mlsysauthorlist}

\mlsysaffiliation{umbc}{Department of Information Systems, University of Maryland, Baltimore County, Baltimore, MD, USA}
\mlsyscorrespondingauthor{Mostafa Cham}{mcham2@umbc.edu}

\mlsyskeywords{Symbolic Regression, Large Language Models, Uncertainty Quantification, Conformal Prediction}

\vskip 0.3in

\begin{abstract}

Large language models (LLMs) can help discover mathematical equations from data, but most LLM-guided symbolic regression methods focus on a single best equation without quantifying uncertainty over alternative functional forms. This is challenging because distinct equations can fit the observed data similarly while producing substantially different extrapolations. Our experiments on the LLM-SRBench dataset show that standard conformal intervals achieve nominal in-distribution coverage but frequently fail under extrapolation. To tackle this challenge, we introduce LLM-SR-UQ, a framework that retains competing equation structures and uses their disagreement to characterize uncertainty about the underlying functional form. We extend the standard fixed-width conformal approach to quantify structural intervals, which measure disagreement among candidate equations and identify where prediction intervals should widen. Extensive experiments show our framework substantially improves coverage attainment by incorporating structural disagreement while preserving in-distribution coverage. Our approach also shows good generalizability when it is applied to other symbolic-regression methods. The code for this paper is available at \url{https://github.com/MostafaCham/LLM-SR-UQ}

\end{abstract}
]

\printAffiliationsAndNotice{}  

\section{Introduction}
\label{sec:intro}

Symbolic regression aims to discover interpretable mathematical equations that describe observed data and, ideally, remain valid beyond the observed range. Recent advances in large language models (LLMs) have improved symbolic regression by enabling the generation and refinement of candidate equations using scientific context and iterative feedback \citep{shojaee2024llmsr,grayeli2024concept,wang2025drsr,srscientist2025}. Despite these advances, current LLM-guided symbolic regression systems typically focus on identifying a single best equation, leaving an important question unanswered: when several equations can explain the observed data, how certain are we that the selected equation is the appropriate functional form, and how reliable are its predictions outside the observed range? 

Different mathematical equations can fit the observed data equally well while representing different hypotheses about the underlying relationship. These differences may be difficult to detect within the observed range but can lead to substantially different predictions under extrapolation. As a result, selecting only the highest-scoring equation can hide whether it is uniquely supported by the data or is one of several plausible alternatives. Classical symbolic regression has explored this problem through Bayesian model selection, ensembles, and other approaches that retain information about competing equations \citep{guimera2020bayesian,fasel2022ensemble,hirsh2022sparsifying,bartlett2024exhaustive}.  However, quantifying such structural uncertainty remains relatively unexplored in LLM-guided symbolic regression.

Evaluating the accuracy of a discovered equation is not sufficient to establish the reliability of its uncertainty estimates. Existing symbolic regression benchmarks primarily assess point-prediction accuracy and symbolic recovery \citep{lacava2021srbench,shojaee2025llmsrbench}. but these metrics do not indicate whether prediction intervals contain the true outcomes at the expected frequency. In particular, an equation may achieve low prediction error within the observed range while producing unreliable uncertainty estimates under extrapolation. For scientific applications, it is therefore important to evaluate not only predictive accuracy but also whether uncertainty estimates remain calibrated under the conditions in which the equation is applied.

Conformal prediction offers a practical approach to estimating predictive uncertainty by constructing prediction intervals from observed prediction errors \citep{vovk2005algorithmic,lei2018distribution,angelopoulos2021gentle}. It provides finite-sample marginal coverage guarantees when calibration and test data are exchangeable, even if the underlying model is incorrect. However, when applied to a single equation selected through symbolic regression, conventional conformal prediction does not explicitly account for alternative equations discarded during the search. This limitation is particularly important under extrapolation, where equations that fit the observed data similarly may produce substantially different predictions. Even when the coefficients of a selected equation are well estimated, uncertainty about its functional form may remain. Once competing equations have been discarded, their disagreement can no longer inform the uncertainty estimate. Consequently, post-hoc calibration may produce reliable intervals within the observed distribution while failing to capture structural ambiguity under extrapolation.

To address these limitations,  we introduce \textsc{LLM-SR-UQ}, which retains competing equation structures throughout the search rather than selecting a single best equation. Evidence-weighted model averaging produces a point predictor, while disagreement among retained equations determines how uncertainty varies across inputs. We extend fixed-width split-conformal intervals by normalizing held-out calibration errors using this structural scale and applying the resulting conformal quantile to construct prediction intervals. These structural intervals widen where candidate predictions diverge and remain narrower where they agree.

We also test whether uncertainty-related feedback improves equation discovery. An optional LLM critic examines fitting diagnostics and coefficient identifiability, with its extrapolation-aware setting additionally using disagreement among candidates on unlabeled inputs beyond the training range. The critic guides proposals without accessing shifted test data or altering retained candidates. Ablations compare it with rule-based feedback.

Our main contributions are summarized as follows: (1) Structural uncertainty framework. We introduce \textsc{LLM-SR-UQ}, a framework that retains multiple candidate equation structures during symbolic regression and combines their disagreement with conformal calibration to construct input-dependent prediction intervals that reflect ambiguity in the underlying functional form (Section~\ref{sec:method}). (2) Evaluation of uncertainty under extrapolation. We systematically evaluate prediction accuracy, coverage, and interval sharpness under both in-distribution (ID) and extrapolative settings. Our experiments demonstrate that incorporating structural disagreement improves empirical coverage under extrapolation, including when applied to independent symbolic-regression methods (Sections \ref{sec:method} and \ref{sec:results}). (3) Calibration-aware search feedback. We investigate whether uncertainty-related information can improve equation discovery through controlled ablations of rule-based feedback and an LLM-based critic. Our results show that the critic does not improve calibration or equation recovery beyond rule-based feedback, highlighting the importance of retaining structural alternatives rather than introducing additional LLM-guided search mechanisms (Section~\ref{sec:method} and Appendix~\ref{app:critic-results}).

\section{Related work}
\label{sec:related}

\textbf{LLM-guided symbolic regression.}
Recent advances in LLM-guided symbolic regression have focused on improving how mathematical equations are generated and refined during search. LLM-SR \citep{shojaee2024llmsr} introduced LLM-driven program synthesis for equation discovery, with performance evaluated through prediction accuracy and equation recovery. Subsequent studies have explored different search strategies, including reusable concept libraries \citep{grayeli2024concept}, in-context iteration \citep{merler2024incontext}, reasoning over data and previous experience \citep{wang2025drsr}, tool-using agents \citep{srscientist2025,thinklike2026}, tree search \citep{igsr2026}, structure-guided evolution \citep{functionevolve2026}, and hierarchical agent systems \citep{asr2026}.
Other studies have examined the role of LLMs in this process, suggesting that improvements may arise primarily from effective search control \citet{lmguide2026}, while selecting equations from a fixed dictionary can sometimes outperform LLM-based evolution \citet{dictionaries2026}. Despite these advances, the emphasis remains largely on improving equation discovery and predictive accuracy rather than quantifying uncertainty arising from alternative functional forms.

Recent studies have addressed the reliability of LLM-guided symbolic regression by improving equation robustness. STRIDE combines critic–executor feedback with diversity-preserving memory to improve out-of-distribution (OOD) robustness \citep{su2026stride}. LLM-PDESR addresses observation noise and memorization by evaluating candidate equations through a Pareto frontier \citep{llmpdesr2026}. However, improving equation robustness does not establish whether the resulting prediction intervals achieve reliable coverage under extrapolation.

\textbf{Structural uncertainty in classical symbolic regression.}
Classical symbolic regression has long explored methods for representing uncertainty over alternative equation structures. Ensemble-SINDy \citep{fasel2022ensemble} uses bootstrap sampling to estimate the inclusion probabilities of candidate terms, while sparsifying-prior model discovery \citep{hirsh2022sparsifying} uses Bayesian priors to quantify uncertainty over model terms. Other approaches, such as the Bayesian machine scientist \citep{guimera2020bayesian} and exhaustive symbolic regression \citep{bartlett2024exhaustive}, explore alternative mathematical expressions through Bayesian sampling and minimum description length, respectively. These methods characterize uncertainty through quantities such as term-inclusion probabilities, distributions over candidate expressions, or relative model scores, rather than necessarily producing prediction intervals with a specified coverage target. Building on this perspective, our work investigates how disagreement among retained equations can be used to construct calibrated prediction intervals in LLM-guided symbolic regression, particularly under extrapolation.

\textbf{Distribution-free calibration.}
Conformal prediction constructs prediction intervals with finite-sample marginal coverage guarantees when calibration and test data are exchangeable \citep{vovk2005algorithmic,lei2018distribution,
angelopoulos2021gentle}. Weighted conformal prediction extends this approach to certain forms of covariate shift by accounting for differences between the calibration and test distributions \citep{tibshirani2019conformal}. In symbolic regression, Conformal-SINDy \citep{conformalsindy2025} applies conformal calibration to predictions from SINDy models. However, to our knowledge, conformal calibration has not been systematically evaluated for LLM-guided symbolic regression under extrapolation, where standard coverage guarantees no longer apply. Existing approaches characterize uncertainty over candidate structures or calibrate predictions from fitted models, but these objectives are typically studied separately. Our work combines disagreement among retained symbolic equations with conformal calibration to construct input-dependent prediction intervals and evaluates their empirical coverage under extrapolation.

\section{Methodology}
\label{sec:method}
 
\subsection{\textbf{Problem setup}}
\label{sec:setup}
 
Given a dataset of observations $\{(x_i,y_i)\}_{i=1}^{N}$, where $x_i \in \mathbb{R}^d$ represents the input variables and $y_i$ is a scalar target, together with brief descriptions of each variable, the goal is to discover a closed-form equation describing their relationship, and uncertainty measurements under both ID and OOD. An equation is
characterized by its \emph{structure} $M$, which defines its functional form, and its \emph{coefficients} $\theta = (c_0,c_1,\ldots)$, which determine the numerical values within that form. For example, $M(x;\theta) = c_0 x_1 e^{-c_1 x_2}$ represents a structure with two unknown coefficients, $c_0$ and $c_1$. For each candidate structure, these coefficients are estimated by minimizing the squared prediction error on the training data. The greater challenge lies in selecting the appropriate structure, since different functional forms may fit the observations similarly well but behave differently under extrapolation. Expressions differing only in coefficient names are considered members of the same structural family. The objective is to obtain both a point predictor $\hat{y}(x)$ and a prediction interval $C_\alpha(x)$ that reflects uncertainty arising from competing equation structures.
 
\subsection{\textbf{Dataset}}
\label{sec:data}
We use the LLM-SRBench dataset~\cite{shojaee2025llmsrbench} and organize the data into separate subsets for equation fitting, uncertainty calibration, and evaluation (Table~\ref{tab:splits}). For each problem, the training data are randomly divided into $\Dfit$ ($80\%$), used to fit and score candidate equations, and $\Dcid$ ($20\%$), used to calibrate prediction intervals. We evaluate the resulting models on two benchmark test sets: $D_{\mathrm{test}}$, containing inputs within the training range (interpolation), and $D_{\mathrm{OOD}}$, containing inputs beyond that range (extrapolation). Both test sets are reserved for final evaluation. In addition, we construct an unlabeled extrapolation region, $D_{\mathrm{cal\text{-}EXT}}$, just beyond the training range to examine how candidate equations behave outside the observed data. This region provides information about disagreement among equations without revealing the true target values. Thus, no extrapolation labels are used during equation search or uncertainty calibration.
 
\subsection{\textbf{Overview}}
\label{sec:overview}
 
\begin{figure*}[t]
\centering
\includegraphics[width=\textwidth]{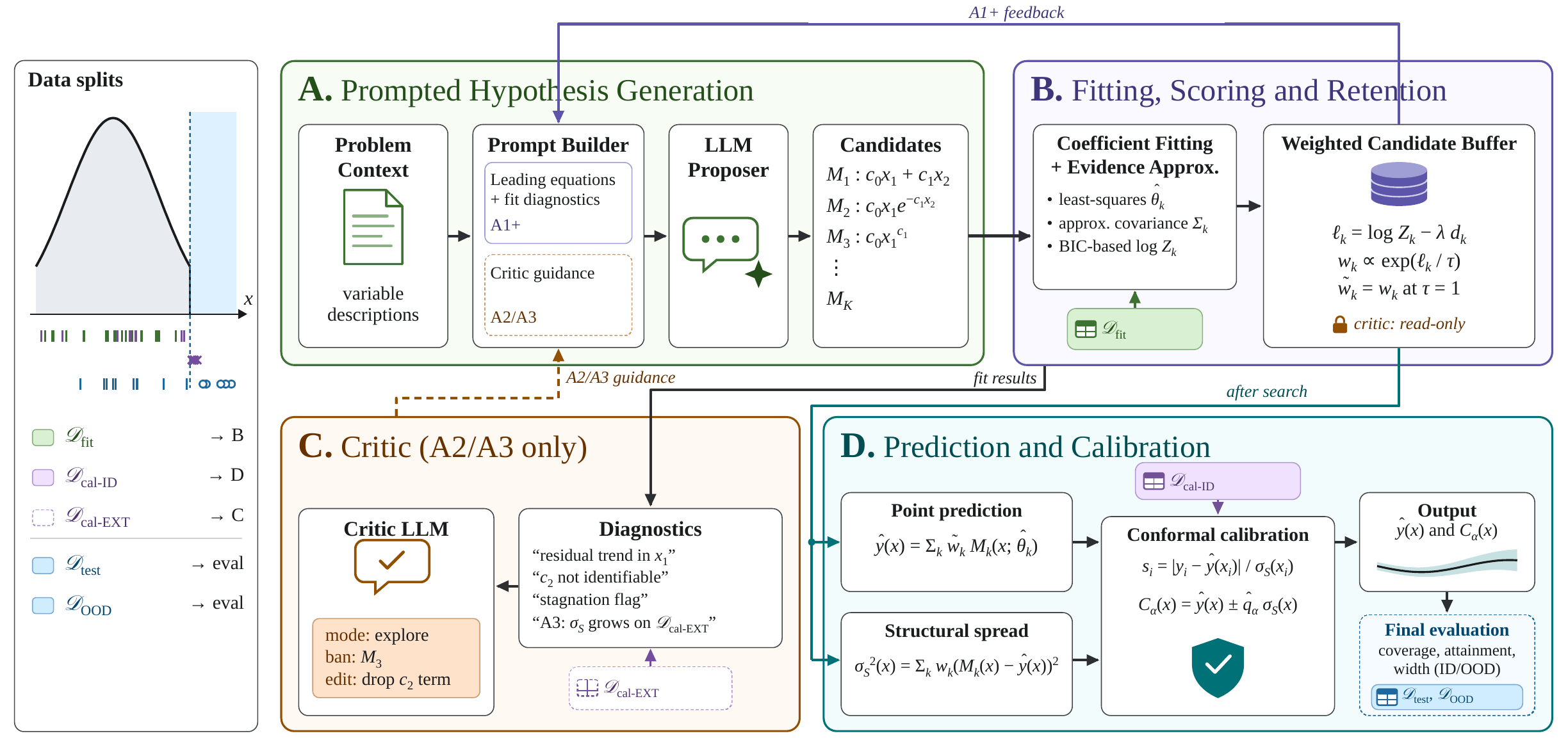}
\vspace{-2em}
\caption{Overview of LLM-SR-UQ. Candidate equations are proposed (A), fitted and retained in a weighted buffer (B). An optional critic guides subsequent proposals without modifying the buffer (C). The retained equations are combined for prediction, while their disagreement informs conformally calibrated structural intervals (D).}
\label{fig:arch}
\end{figure*}
 
LLM-SR-UQ consists of an iterative equation search followed by prediction and uncertainty calibration, organized into four stages (Figure~\ref{fig:arch}). In \textbf{Stage A}, an LLM proposes candidate equation structures. In \textbf{Stage B}, each candidate is fitted and scored, and the highest-scoring structures are retained in a buffer with associated weights. In \textbf{Stage C}, when enabled, a second LLM acts as a critic, examining numerical diagnostics of the retained candidates and providing feedback to guide subsequent proposals. Finally, in \textbf{Stage D}, the retained equations are combined to form a single predictor, while their disagreement determines how prediction intervals vary across the input space. Conformal calibration then adjusts the overall interval width based on observed prediction errors. To measure what each component contributes, we switch Stage C on in four steps, which we call \emph{rungs}: \textbf{A0} gives the proposer only the problem description; \textbf{A1} adds a fixed, rule-based summary of the current fits; \textbf{A2} adds the critic; and \textbf{A3} additionally shows the critic label-free evidence about extrapolation from $\Dcext$. All rungs use the same budget of LLM calls (Section~\ref{sec:ladder}).
 
\subsection{\textbf{Stage A: proposing structures}}
\label{sec:propose}
 
At each proposal round, the LLM receives the problem description and information about the input variables, and generates $K=8$ candidate equation structures with free coefficients $c_0,c_1,\ldots$. Starting with configuration A1, the proposer also receives a fixed summary of the current search results, including the leading equations and their fitting errors, poorly identified coefficients, unused input variables, and systematic residual patterns. In configurations A2 and A3, the latest critic guidance is additionally provided to help refine subsequent proposals. The proposer does not directly access the numerical training data. The feedback rules and prompt templates are provided in Appendices~\ref{app:impl} and~\ref{app:prompts}.
 
\subsection{\textbf{Stage B: scoring and retaining structures}}
\label{sec:bayes}
 
Each proposed equation structure $M_k$ is evaluated by fitting its coefficients on $D_{\mathrm{fit}}$ using least squares. This produces the estimated coefficients $\hat{\theta}_k$, from which we calculate the residual sum of squares ($\mathrm{RSS}_k$) and the estimated noise variance $\hat{\sigma}_k^2$:
\begin{equation}
\mathrm{RSS}_k =
\sum_{i \in D_{\mathrm{fit}}}
\left(y_i - M_k(x_i;\hat{\theta}_k)\right)^2,
\quad
\hat{\sigma}_k^2 = \frac{\mathrm{RSS}_k}{n},
\label{eq:residual_variance}
\end{equation}
where $n$ is the number of observations in $D_{\mathrm{fit}}$. Although more flexible equations may achieve lower training error, they can also overfit the observed data. We therefore evaluate candidate structures using an approximation to their Bayesian evidence, which accounts for both fitting quality and model complexity.

To compare candidate structures, we use the Bayesian Information Criterion (BIC), which balances fitting accuracy against model complexity \citep{mackay1992bayesian}. For each structure $M_k$, we compute
\begin{equation}
\mathrm{BIC}_k =
n\log\hat{\sigma}_k^2 + d_k\log n,
\qquad
\log Z_k \approx -\frac{1}{2}\mathrm{BIC}_k,
\label{eq:bic}
\end{equation}

where $d_k$ is the number of free coefficients. A lower residual variance improves the BIC score, while additional coefficients increase the complexity penalty.

We use BIC to approximate the log Bayesian evidence $\log Z_k$, which measures how well a structure explains the observed data after accounting for uncertainty in its coefficients \citep{mackay1992bayesian}. Finally, each candidate receives the score
\begin{equation}
\ell_k = \log Z_k - \lambda d_k,
\label{eq:model_score}
\end{equation}
where $\lambda = 0.05$ introduces a small additional penalty for model complexity, favoring simpler structures when their evidence is similar.

The buffer retains up to $B=24$ highest-scoring candidate equations. Duplicate equation structures are removed, but distinct structures belonging to the same structural family may coexist in the buffer. However, directly assigning weights according to Bayesian evidence can cause nearly all the weight to concentrate on a single equation, particularly when the dataset is large. This limits the contribution of alternative structures, even when their predictions differ substantially under extrapolation. To preserve their influence, we introduce temperature-controlled weights:
\begin{equation}
w_k =
\frac{\exp(\ell_k/\tau)}
{\sum_{j=1}^{B}\exp(\ell_j/\tau)},
\label{eq:tempered_weights}
\end{equation}
where $\tau > 0$ controls how strongly differences in model scores affect the weights. At $\tau=1$, the weights are proportional to the Bayesian evidence multiplied by the structural prior; we denote these untempered weights by $\tilde{w}_k$. Increasing $\tau$ distributes weight more evenly among the retained equations.

Because score differences depend on the dataset size, a fixed temperature may produce different levels of weight concentration across problems. We therefore adjust $\tau$ to control the effective number of structures contributing to the weighted ensemble, using the measure introduced by~\citep{kish1965survey}:
\begin{equation}
B_{\mathrm{eff}} =
\left(\sum_{k=1}^{B}w_k^2\right)^{-1}.
\label{eq:effective_models}
\end{equation}
Here, $w_k$ is the tempered weight of the $k$-th retained structure, and $B$ is the number of retained structures. The effective model count $B_{\mathrm{eff}}$ ranges from 1, when a single structure dominates, to $B$, when all structures receive equal weight. We select $\tau$ to achieve a target of $B_{\mathrm{eff}}=0.35B\approx8.4$, allowing multiple structures to contribute to the uncertainty estimate. Our sensitivity analysis shows that the results remain stable across a broad range of target effective model counts, provided $B_{\mathrm{eff}}>1$ (Appendix~\ref{app:posterior}).

In our experiments, the untempered weights concentrate almost entirely on the highest-scoring equation, with a median effective model count of 1.00 across configurations. Consequently, evidence-proportional model averaging provides little structural disagreement. We therefore use tempered weights to measure disagreement among the retained equations, treating this spread as an empirical measure of structural uncertainty rather than a calibrated Bayesian posterior.

\subsection{\textbf{Stage C: the critic}}
\label{sec:critic}
 
The critic examines the current search results and provides feedback to guide subsequent equation proposals. Every fifth language-model call is allocated to the critic instead of the proposer. The critic receives numerical diagnostics of the retained candidates, including residual patterns, poorly identified coefficients, repeatedly proposed structural families, and signs of search stagnation. In configuration A3, it additionally receives label-free information from $D_{\mathrm{cal\text{-}EXT}}$, describing disagreement among the three leading equations and changes in prediction-interval width beyond the observed range.

Based on these diagnostics, the critic recommends structures to avoid, alternative directions to explore, and up to three modifications of existing candidates. Each suggested modification must be supported by the available diagnostics and validated before use. Importantly, the critic only influences subsequent proposals; candidate retention remains determined by the score $\ell_k$. The critic's output format and validation rules are provided in Appendices~\ref{app:impl} and~\ref{app:prompts}.

\subsection{\textbf{Stage D: Structural Intervals and Conformal Calibration}}
\label{sec:calibration}
After completing the search, we combine the retained equations into a single predictor using Bayesian model averaging (BMA) \citep{hoeting1999bma}. The prediction at input $x$ is computed as the weighted average:
\begin{equation}
\hat{y}(x) =
\sum_{k=1}^{B}
\tilde{w}_k M_k(x;\hat{\theta}_k),
\label{eq:bma_prediction}
\end{equation}
where $M_k(x;\hat{\theta}_k)$ is the prediction of the $k$-th fitted structure and $\tilde{w}_k$ is its untempered evidence weight. Because these weights are strongly concentrated on the highest-scoring structure, the resulting prediction is typically close to that of the single best equation. 

To measure structural disagreement, we examine how much the retained equations differ from the final prediction at each input $x$. We define the structural spread using the tempered weights as
\begin{equation}
\sigma_S^2(x) =
\sum_{k=1}^{B}
w_k \left(
M_k(x;\hat{\theta}_k) - \hat{y}(x)
\right)^2.
\label{eq:structural_spread}
\end{equation}
Here, $w_k$ is the tempered weight of the $k$-th retained structure, $M_k(x;\hat{\theta}_k)$ is its prediction, and $\hat{y}(x)$ is the final predictor defined in Equation~\ref{eq:bma_prediction}. The structural spread is small when candidate equations make similar predictions and increases when their predictions diverge. We use untempered weights for the point predictor to emphasize the best-supported equation, while tempered weights allow alternative structures to contribute more substantially to the disagreement measure.
 
Structural disagreement alone does not guarantee that prediction intervals achieve the desired coverage. We therefore use split conformal prediction \citep{vovk2005algorithmic,lei2018distribution} to calibrate the structural spread using the held-out set $\Dcid$, containing $m$ observations. For each calibration point, we compute the normalized prediction error
\begin{equation}
s_i =
\frac{|y_i-\hat{y}(x_i)|}{\sigma_S(x_i)},
\qquad i \in D_{\mathrm{cal\text{-}ID}}.
\label{eq:conformal_score}
\end{equation}
Here, $|y_i-\hat{y}(x_i)|$ is the absolute prediction error, and $\sigma_S(x_i)$ is the structural spread at the same input. The normalized score measures the prediction error relative to the disagreement among candidate equations. We then select the calibration quantile $\hat{q}_\alpha$ as the $\lceil(m+1)(1-\alpha)\rceil$-th smallest score, where $\alpha=0.1$ corresponds to a nominal coverage level of 90\%. The prediction interval for a new input $x$ is defined as
\begin{equation}
C(x) =
\left[
\hat{y}(x)-\hat{q}_\alpha\sigma_S(x),
\hat{y}(x)+\hat{q}_\alpha\sigma_S(x)
\right].
\label{eq:prediction_interval}
\end{equation}
The conformal quantile $\hat{q}_\alpha$ determines the overall scale of the intervals, while $\sigma_S(x)$ adjusts their width according to the disagreement among retained equations. Consequently, intervals expand where candidate predictions diverge and remain narrower where they agree. Under exchangeability between calibration and test observations, this construction provides finite-sample marginal coverage of at least $1-\alpha$, regardless of whether the candidate equations are correctly specified. However, this guarantee does not generally extend to $D_{\mathrm{OOD}}$, where predictions are evaluated beyond the observed input range. We examine empirical coverage under this extrapolation setting in Section~\ref{sec:results}.

To assess the contribution of structural disagreement, we compare two interval constructions centered on the same predictor $\hat{y}(x)$. The proposed \textbf{structural intervals (S)}, defined in Equation~\ref{eq:prediction_interval}, use the input-dependent structural spread $\sigma_S(x)$ to scale interval widths according to disagreement among retained equations. In contrast, the \textbf{split conformal baseline (split)} uses a constant unit scale, yielding absolute-residual calibration scores $s_i=|y_i-\hat{y}(x_i)|$ and intervals of fixed width across inputs, determined by the conformal quantile.

Both constructions use the same fitting and calibration data and provide finite-sample marginal coverage guarantees under exchangeability. Since they share the same point predictor, their comparison isolates the effect of using structural disagreement to determine interval width. Structural intervals are independent of the algorithm used to discover candidate models. Given multiple fitted candidates, their prediction disagreement provides an input-dependent uncertainty scale that can be calibrated using held-out observations without modifying the underlying search procedure. We demonstrate this transfer with LLM-SR and Ensemble-SINDy in Section~\ref{sec:transfer}. In all cases, we report the highest-weight structure as the discovered equation, while the retained ensemble is used for prediction and uncertainty estimation.

\section{Evaluation and experiment setup}
\label{sec:protocol}

We evaluate prediction accuracy and uncertainty reliability at the problem level, focusing on performance under extrapolation. All prediction intervals target 90\% coverage. Because distribution shifts vary across domains, we report domain-specific results alongside pooled summaries where appropriate.

\subsection{Accuracy}
\label{sec:acc-metrics}

We evaluate prediction accuracy using median normalized mean squared error (NMSE) on both in-distribution and out-of-distribution test sets, along with accuracy-to-tolerance ($\mathrm{Acc}_{0.1}$) and symbolic accuracy. The metric $\mathrm{Acc}_{\tau}$ measures the fraction of problems for which all test predictions have relative errors within a specified tolerance $\tau$. We use $\tau=0.1$ and count non-finite predictions as failures.

To evaluate symbolic accuracy (SA@$k$) we replace numerical constants and free coefficients in the discovered and reference equations with canonical placeholders, then compare the resulting using SymPy. Because different symbolic expressions can represent the same function, we separately evaluate functional recovery after refitting candidate coefficients. A candidate is recovered if its predictions match the reference within specified tolerances across training, in-distribution, out-of-distribution, and additional extrapolation points. We report functional recovery at ranks 1 and 3, alongside SA@$k$ for the highest-weight candidate structures. The numerical thresholds and detailed recovery criteria are provided in Appendix~\ref{app:impl}.

\subsection{Uncertainty}
\label{sec:eval-uncertainty}
\label{sec:uq-metrics}

We evaluate predictive uncertainty using coverage attainment, interval sharpness, and interval score. Coverage alone is insufficient, since wider prediction intervals can achieve higher coverage without providing useful precision. We therefore assess both how often intervals contain the true outcomes and how narrow they remain.

\textbf{Coverage attainment.} We define the attainment rate as
\begin{equation}
A_\alpha = \frac{1}{N}\sum_{p=1}^{N}
\mathbf{1}\left[
\operatorname{cov}^{\mathrm{OOD}}_p
\geq 1-\alpha-\epsilon
\right],
\label{eq:attainment}
\end{equation}

where $N$ is the number of benchmark problems and $\operatorname{cov}^{\mathrm{OOD}}_p$ is the empirical coverage on the shifted test set for problem $p$. With a nominal coverage level of $1-\alpha=0.90$, the tolerance $\epsilon=0.02$ allows problems with at least 88\% coverage to count as attaining the target.

\textbf{Sharpness at coverage.} To assess interval width among successfully covered problems, we define
\begin{equation}
S_\alpha =
\operatorname*{median}_{
p:\operatorname{cov}^{\mathrm{OOD}}_p
\geq 1-\alpha-\epsilon}
\frac{\overline{|C_p|}}
{\operatorname{sd}(y_p^{\mathrm{train}})}.
\label{eq:sharpness}
\end{equation}

Here, $\overline{|C_p|}$ is the mean interval width for problem $p$, normalized by the standard deviation of its training targets. Lower values indicate narrower intervals, but $S_\alpha$ is computed only over problems meeting the coverage threshold and must therefore be interpreted alongside the attainment rate.

\textbf{Interval score.} To jointly evaluate interval width and coverage errors, we use the interval score \citep{gneiting2007strictly}:
\begin{equation}
\mathrm{IS}_{\alpha}(C,y)
= (u-\ell)
+ \frac{2}{\alpha}(\ell-y)_+
+ \frac{2}{\alpha}(y-u)_+,
\label{eq:interval_score}
\end{equation}

where $\ell$ and $u$ are the lower and upper bounds of prediction interval $C$, $y$ is the observed target, and $(z)_+=\max(z,0)$. The score penalizes both wider intervals and observations falling outside their bounds, with lower values indicating better performance.

We additionally report ID coverage, absolute calibration error, OOD coverage distributions, Wilson confidence intervals for attainment rates, and interval-width ratios between ID and OOD. Search diversity is measured using distinct structural families and repeated-family rates rather than the effective model count, which is controlled by temperature.

\subsection{Simplicity and cost}
\label{sec:simplicity}

We measure equation complexity using expression-tree node counts and free-parameter counts, and computational cost using language-model calls and estimated cost per problem. We additionally report a plateau statistic measuring the fraction of proposer rounds after the last improvement in the best evidence score. Detailed metric definitions and token-usage limitations are provided in Appendix~\ref{app:impl}.


\subsection{Experimental setup}
\label{sec:setup-exp}
\label{sec:ladder}

\textbf{Benchmark and splits.}
We evaluate on the LSR-Synth subset of LLM-SRBench \citep{shojaee2025llmsrbench}, comprising 129 problems across reaction kinetics, population dynamics, stress–strain, and physical oscillators. Each problem provides 4,000 training observations and separate 500-point test sets for interpolation and extrapolation beyond the training range. We exclude LSR-Transform because it lacks an extrapolation split. Data partitions and the unlabeled extrapolation region are described in Section~\ref{sec:data} and Appendix~\ref{app:impl}. Each problem has a three-hour wall-clock limit, with timeouts recorded as failures and excluded from paired comparisons.

\textbf{Development discipline.}
We reserve 20 problems (five per domain) for prompt development and use the remaining 109 for held-out evaluation. All prompts are frozen before final evaluation, and the critic prompt's fingerprint is recorded. Each held-out problem is evaluated once per configuration, with paired comparisons restricted to problems completed by both configurations.

\textbf{The ablation ladder.}
To assess the contribution of search feedback, we compare configurations A0--A3 (Section~\ref{sec:overview}) under a fixed budget of 125 LLM calls per problem. Adjacent comparisons evaluate the effects of rule-based diagnostics (A1 vs.\ A0), LLM critic guidance (A2 vs.\ A1), and additional label-free extrapolation information (A3 vs.\ A2). In A2 and A3, 24 calls are allocated to the critic, leaving 101 proposer calls, compared with 125 proposer calls in A0 and A1. Thus, all configurations have the same total LLM-call budget, although the number of equation-proposal rounds differs. The complete ablation configurations and their corresponding experimental comparisons are detailed in Appendix~\ref{app:critic-results}, Table~\ref{tab:ladder}.

\subsection{Baselines}
\label{sec:baselines}
We compare LLM-SR-UQ with three complementary baselines under a common evaluation protocol. For LLM-based comparisons, we use GPT-4o-mini and Llama-3.1-8B-Instruct.

The first baseline is \textbf{LLM-SR} \citep{shojaee2024llmsr}, and we use its official implementation and published 1k-sample configuration. We compare its single best equation with split-conformal intervals against the top 24 distinct equations from its experience buffer, which we refit and evidence-weight using our procedure. Applying both split-conformal and structural intervals to the same top-24 predictor isolates the effect of structural disagreement without changing the underlying search.

\textbf{Direct prompting.} Each LLM generates a single equation without iterative search or diagnostic feedback. We apply split-conformal intervals because no alternative equation structures are retained. This baseline tests the benefit of iterative equation search relative to direct generation.

\textbf{Ensemble-SINDy.} As a non-LLM baseline, we use Ensemble-SINDy~\citep{fasel2022ensemble}, which generates multiple sparse-regression models from a shared candidate library. We apply split-conformal and structural intervals to its ensemble to test transfer beyond LLM-guided search. This baseline is relevant because LSR-Synth targets are the right-hand sides of ordinary differential equations, which describe how dynamical systems evolve over time. SINDy is specifically designed to recover such governing expressions from data.

All baselines evaluated in our experiments use the same fitting set $D_{\mathrm{fit}}$ and calibration set $D_{\mathrm{cal\text{-}ID}}$ as LLM-SR-UQ. Because only 80\% of the available training data are used for fitting, their results may differ from those originally reported.

\begin{figure*}[t]
\centering
\includegraphics[width=\textwidth]{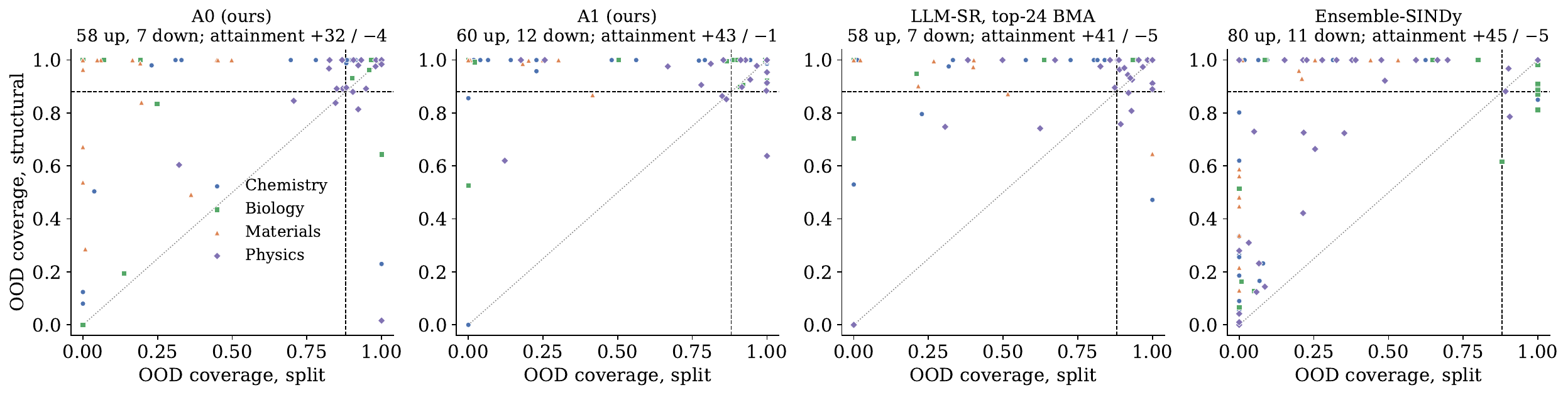}
\vspace{-2em}
\caption{Problem-level OOD coverage under split-conformal and structural intervals for A0, A1, LLM SR top-24 BMA, and Ensemble-SINDy. Each point represents a held-out problem, with split-conformal coverage on the horizontal axis and structural coverage on the vertical axis. The diagonal indicates equal coverage. Dashed lines mark the attainment threshold of 0.88, corresponding to the nominal target of 0.90 with tolerance $\epsilon=0.02$. Panel headings report counts of coverage increases and decreases and transitions into or out of attainment.}
\label{fig:ood_coverage_pairs}
\end{figure*}

\section{Results}
\label{sec:results}

 Unless stated otherwise, results are reported on 109 held-out problems using one search seed, with paired comparisons restricted to problems completed by both methods.

\subsection{Structural disagreement improves uncertainty under shift}
\label{sec:calib-results}

Table~\ref{tab:main_results} compares structural and split-conformal intervals on held-out problems using identical point predictors, fitting data, and calibration data. Both maintain approximately 90\% in-distribution coverage, but their performance differs substantially under extrapolation. Structural intervals increase OOD coverage attainment from 0.58 to 0.83 for A0 and from 0.53 to 0.92 for A1. Because both constructions use identical point predictions, these improvements arise from incorporating structural disagreement into the prediction intervals rather than from improved predictive accuracy.

Figure~\ref{fig:ood_coverage_pairs} compares problem-level OOD coverage under split-conformal and structural intervals for A0, A1, LLM-SR top-24, and Ensemble-SINDy. Structural intervals increase coverage on 58 problems at A0 and 60 at A1, compared with decreases on 7 and 12 problems, respectively. At the coverage-attainment threshold, 32 and 43 problems move from failure to attainment, while only 4 and 1 move in the opposite direction. The same pattern extends to independent search methods, with 41 favorable versus 5 unfavorable attainment transitions for LLM-SR and 45 versus 5 for Ensemble-SINDy. These paired comparisons show that the improvements extend across many individual problems and are not limited to our search procedure, although structural intervals do not improve coverage uniformly.

Higher OOD coverage could result from wider intervals rather than more informative uncertainty estimates. Because sharpness $S_{0.1}$ is computed only for problems meeting the coverage-attainment threshold, it cannot be compared directly across methods. We therefore compare OOD interval widths on problems where both constructions attain coverage. The median structural-to-split width ratio is 1.16 at A0 ($p=0.77$) and 0.98 at A1 ($p=0.22$), based on Wilcoxon signed-rank tests. Neither comparison shows a statistically significant difference, although this restricted analysis does not establish equivalent widths across all problems. We additionally use the OOD interval score to evaluate the coverage--width trade-off across the full benchmark.

The OOD interval score provides a complementary evaluation of the trade-off between coverage and interval width. At A0, structural intervals have a slightly higher median interval score than split-conformal intervals ($1.92\times10^{-2}$ versus $1.73\times10^{-2}$, $p=0.19$), indicating worse performance on this metric. At A1, structural intervals achieve a lower score ($2.77\times10^{-3}$ versus $5.66\times10^{-3}$, $p=0.11$). Neither difference is statistically significant. These findings suggest that structural disagreement improves coverage attainment, although its effect on the overall coverage--width trade-off varies across configurations. Illustrative examples of input-dependent interval widening are provided in Appendix~\ref{app:stageD} (Figure~\ref{fig:example}).

 
\subsection{The uncertainty construction transfers across search methods}
\label{sec:transfer}

To assess whether structural disagreement improves uncertainty estimates beyond our search procedure, we apply structural intervals to candidate sets produced by LLM-SR and Ensemble-SINDy. For each method, we compare split-conformal and structural intervals using identical point predictions, fitting data, and in-distribution calibration data. This comparison isolates the effect of the interval construction from differences in the underlying equation search.

\begin{table*}[t]
\centering
\caption{Prediction accuracy and uncertainty performance
across symbolic-regression methods.}
\label{tab:main_results}
\scriptsize
\setlength{\tabcolsep}{2.5pt}
\renewcommand{\arraystretch}{1.12}
\resizebox{0.85\textwidth}{!}{%
\begin{tabular}{@{}llrrr rr rr@{}}
\toprule
& & \multicolumn{2}{c}{NMSE} & \multicolumn{2}{c}{Coverage}
  & \multicolumn{1}{c}{OOD attainment} & \multicolumn{1}{c}{OOD sharpness} & \multicolumn{1}{c}{OOD IS} \\
\cmidrule(lr){3-4}\cmidrule(lr){5-6}
Method & Interval & ID & OOD & ID & OOD
  & $A_{0.1}$ & $S_{0.1}$ & IS \\
\midrule
Direct, Llama-8B     & split & 2.1e-2 & 2.84 & 0.902 & 0.629 & 0.55 (60/109) & 4.45e-1 & 9.08e-1 \\
Direct, GPT-4o-mini & split & 2.0e-2 & 2.76 & 0.894 & 0.646 & 0.57 (62/109) & 3.83e-1 & 7.94e-1 \\
\midrule
Ensemble-SINDy & split      & 3.2e-7 & 6.3e-2 & 0.899 & 0.365 & 0.26 (28/109) & \textbf{2.03e-4} & 5.58e-2 \\
Ensemble-SINDy & structural & \underline{3.2e-7} & 6.3e-2 & 0.900 & 0.762 & 0.62 (68/109) & 9.29e-3 & 1.89e-2 \\
\midrule
LLM-SR, single best & split      & 4.2e-6 & 2.2e-3 & 0.891 & 0.627 & 0.52 (57/109) & 4.82e-3 & 1.54e-2 \\
LLM-SR, top-24      & split      & 4.2e-6 & 2.2e-3 & 0.883 & 0.635 & 0.53 (58/109) & 4.80e-3 & 1.54e-2 \\
LLM-SR, top-24      & structural & 4.2e-6 & \underline{2.2e-3} & 0.891 & \underline{0.928} & \underline{0.86} (94/109) & 7.28e-3 & 8.26e-3 \\
\midrule
Ours, A0 & split      & 3.3e-6 & 2.4e-3 & 0.902 & 0.671 & 0.58 (63/109) & 7.68e-3 & 1.73e-2 \\
Ours, A0 & structural & 3.3e-6 & 2.4e-3 & 0.902 & 0.906 & 0.83 (91/109) & 1.27e-2 & 1.92e-2 \\
Ours, A1 & split      & 2.0e-8 & 6.0e-5 & 0.901 & 0.627 & 0.53 (58/109) & \underline{2.32e-4} & \underline{5.66e-3} \\
Ours, A1 & structural & \textbf{2.0e-8} & \textbf{6.0e-5} & 0.902 & \textbf{0.964} & \textbf{0.92 (100/109)} & 1.12e-3 & \textbf{2.77e-3} \\
\bottomrule
\end{tabular}%
}

\end{table*}

Table~\ref{tab:main_results} compares prediction accuracy and uncertainty performance across all 109 held-out problems. The direct-prompting baselines produce a single equation per problem and are therefore evaluated only with split-conformal intervals. In contrast, Ensemble-SINDy and the top-24 LLM-SR variant retain multiple candidate equations, allowing both split-conformal and structural intervals to be evaluated. We also include LLM-SR's single-best variant to distinguish the effect of retaining multiple equations from the effect of using their structural disagreement. Within each method, both interval constructions use identical point predictions, as reflected in their unchanged NMSE values. The comparison therefore evaluates whether structural disagreement improves coverage and attainment under extrapolation, while sharpness and interval score characterize the associated trade-off in interval width.

The LLM-SR results provide a controlled comparison of retaining multiple equations and using their structural disagreement to construct uncertainty intervals. With the single-best LLM-SR equation, split-conformal intervals attain the OOD coverage target on 57 of 109 problems (0.52). Retaining the top 24 equations while using split conformal intervals produces only a marginal improvement, reaching 58 problems (0.53). In contrast, applying structural intervals to the same top-24 candidate set increases OOD coverage from 0.635 to 0.928 and coverage attainment from 0.53 to 0.86 (94/109 problems). The median OOD interval score also decreases from $1.54\times10^{-2}$ to $8.26\times10^{-3}$, while the point-prediction NMSE remains unchanged. These results show that the improvement arises from incorporating disagreement among candidate equations into the uncertainty intervals, rather than simply retaining multiple equations.

Ensemble-SINDy provides a complementary test because it generates candidate equations through sparse regression rather than LLM-guided search. Structural intervals increase OOD coverage from 0.365 to 0.762 and coverage attainment from 0.26 (28/109 problems) to 0.62 (68/109 problems). In-distribution coverage remains approximately 0.90, and both ID and OOD NMSE remain unchanged because the point predictions are identical. The median OOD interval score also decreases from $5.58\times10^{-2}$ to $1.89\times10^{-2}$, indicating a better observed coverage--width trade-off. Although the reported sharpness statistic increases from $2.03\times10^{-4}$ to $9.29\times10^{-3}$, these values are computed over different sets of problems meeting the coverage-attainment threshold and are not directly comparable. Together, these results show that structural disagreement can improve extrapolation coverage even when the candidate equations come from a non-LLM search method.

\begin{figure}[t]
\centering
\includegraphics[width=\columnwidth]{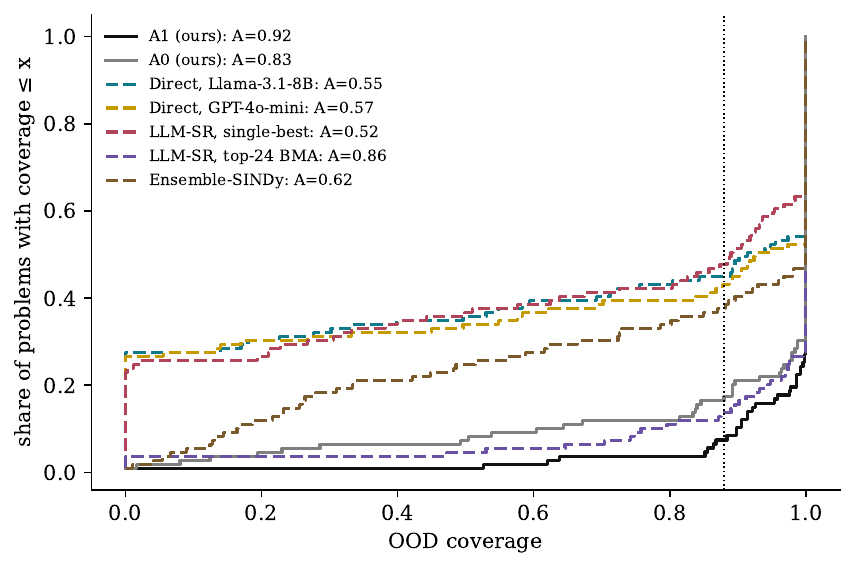}
\vspace{-2em}
\caption{Empirical cumulative distribution of OOD coverage across 109 held-out problems. Each curve shows the fraction of problems with empirical coverage at or below the horizontal-axis value. A0, A1, LLM-SR top-24 BMA, and Ensemble-SINDy use structural intervals. Direct-prompting baselines and LLM-SR single-best use split-conformal intervals. The vertical dotted line marks the attainment threshold of 0.88. Legend values report coverage attainment $A_{0.1}$.}
\label{fig:ood_coverage_ecdf}
\end{figure}

Figure~\ref{fig:ood_coverage_ecdf} provides a complementary view of uncertainty performance by showing the distribution of OOD coverage across held-out problems. Each curve represents the fraction of problems with empirical coverage at or below a given value. Near the 0.88 attainment threshold, lower curves indicate fewer coverage failures. A1 achieves the highest attainment rate of 0.92, followed by LLM-SR top-24 with structural intervals at 0.86 and A0 at 0.83. Ensemble-SINDy with structural intervals also outperforms the single-equation baselines in attainment. The coverage distributions reveal substantial differences across methods and show that none achieves the attainment target on every held-out problem.

The cross-method comparison also places the complete LLM-SR-UQ system in context. With structural intervals, A0 achieves OOD coverage attainment of 0.83 (91/109 problems), slightly below LLM-SR top-24 at 0.86 (94/109), but above Ensemble-SINDy at 0.62 (68/109). A1 increases attainment to 0.92 (100/109), the highest among the evaluated methods. It also achieves the lowest median ID and OOD NMSE, at $2.0\times10^{-8}$ and $6.0\times10^{-5}$, respectively, together with the lowest OOD interval score of $2.77\times10^{-3}$. These results demonstrate that A1 combines strong point-prediction accuracy with improved uncertainty performance under extrapolation.

The difference between A0 and A1 highlights the contribution of mechanical feedback to equation discovery. Unlike A0, A1 provides the proposer with fitting diagnostics and information about previously evaluated candidate equations, allowing subsequent proposals to respond to observed errors. This reduces median ID NMSE from $3.3\times10^{-6}$ to $2.0\times10^{-8}$ and median OOD NMSE from $2.4\times10^{-3}$ to $6.0\times10^{-5}$. These improvements indicate that mechanical feedback helps the search identify more accurate equations. Structural intervals serve a complementary purpose by using disagreement among the retained equations to characterize uncertainty under extrapolation. The LLM-SR and Ensemble-SINDy results further demonstrate that this uncertainty construction can improve coverage independently of the proposed search procedure.

\textbf{Search diversity and the role of the critic.} Search diversity provides further insight into the effect of mechanical feedback. At A1, only 22\% of proposed equations belong to previously explored structural families, compared with 31\% at A0. Under the same budget of approximately 1,000 candidate equations, A1 explores 774 distinct structural families, whereas A0 explores 690. This suggests that mechanical feedback reduces repeated proposals and encourages exploration of new equation structures, which may contribute to the improved predictive accuracy observed at A1. Despite this increased exploration, the discovered equations remain relatively compact, with 6.7--8.2 fitted coefficients and 20--25.5 expression-tree nodes across the evaluated configurations.

The calibration-aware critic produces a different pattern of search behavior. Across critic-assisted configurations, repeated proposals account for 33--39\% of candidates, compared with 22\% at A1, while the number of distinct structural families decreases from 774 to 494--543. However, the critic tends to produce simpler reported equations. Relative to A1, expression-tree size decreases from 25.1 to 22.5 nodes with the 8B critic ($p=0.007$) and to 21.9 nodes with GPT-4o ($p=8\times10^{-5}$) at A3. These results indicate that critic-assisted search yields more compact equations but does not improve uncertainty performance.

The critic ablation indicates that adding a calibration-aware critic does not improve OOD coverage attainment under the fixed budget of 125 language-model calls. Mechanical feedback alone (A1) achieves the highest attainment of 0.92 (100/109 problems), whereas the critic-assisted A2 and A3 configurations attain 0.82–0.88 across both critic models. Although the critic produces valid equation edits, particularly with GPT-4o, these edits rarely outperform the leading candidates under the evidence criterion. Consequently, they have limited influence on the retained equations. Because critic calls replace some proposer calls within the fixed budget, these results characterize the overall critic-assisted search configuration rather than the isolated effect of the critic. The main uncertainty improvements are therefore achieved without the critic, through retaining alternative equations and using their disagreement to construct input-dependent intervals.

\section{Conclusion}
\label{sec:conclusion}
LLM-guided symbolic regression often selects a single equation, overlooking uncertainty among functional forms that fit observed data similarly but diverge under extrapolation. We introduce LLM-SR-UQ, which combines disagreement among retained equations with conformal calibration to construct input-dependent prediction intervals. Experiments on LLM-SRBench show that split-conformal intervals maintain approximately nominal in-distribution coverage but frequently fail under extrapolation, with prediction accuracy alone failing to identify these errors. Across 109 held-out problems, structural intervals improve OOD coverage attainment from 0.58 to 0.83 at A0 and from 0.53 to 0.92 at A1, while preserving in-distribution calibration.

These gains also extend to independent search methods. Structural intervals increase OOD attainment from 0.53 to 0.86 for LLM-SR top-24 and from 0.26 to 0.62 for Ensemble-SINDy without changing point predictions. Within LLM-SR-UQ, mechanical feedback improves equation accuracy and search diversity, whereas the calibration-aware critic produces simpler equations without additional coverage gains under the fixed computational budget. Overall, the results demonstrate the value of structural disagreement for improving empirical uncertainty estimates under extrapolation.

\section*{Acknowledgments}
This research has been funded by the NSF HDR Institute for Harnessing Data and Model Revolution in the Polar Regions (iHARP), Award \#2118285.

\bibliography{references}
\bibliographystyle{mlsys2025}


\newpage
\appendix
\onecolumn





\section*{Appendix}

\section{Implementation details}\label{app:impl}

This appendix collects the implementation choices omitted from
Section~\ref{sec:method} for readability.

\subsection{Data splits}

\begin{table}[h]
\centering\small
\caption{Data splits. ID: in-distribution (inputs inside the training range).
EXT and OOD: extrapolation and out-of-distribution (inputs beyond it).}
\label{tab:splits}
\begin{tabular}{@{}llll@{}}
\toprule
Split & Content & Labels & Role \\
\midrule
$\Dfit$  & $80\%$ of the training rows ($n=3{,}200$) & yes & fit and score structures \\
$\Dcid$  & the other $20\%$ ($m=800$)                & yes & calibrate intervals \\
$\Dcext$ & points just beyond the training range     & no  & measure disagreement (critic only) \\
$\Dtest$ & benchmark test set, inside the range      & yes & final evaluation \\
$\Dood$  & benchmark test set, beyond the range      & yes & final evaluation \\
\bottomrule
\end{tabular}
\end{table}

\subsection{Stage A: prompts and proposals}
In A0, the proposer receives only the problem description at every round. From A1 onward, a fixed-format feedback block, limited to $5,000$ characters, summarizes the three highest-weight structures, their training NMSE and coefficient counts, poorly identified coefficients and their associated terms, unused input variables, systematic residual patterns, protected-operator activations, repeated structural families, and search stagnation. The leading structures are included as search diagnostics rather than separate few-shot examples. In A2 and A3, the prompt also includes the critic's latest guidance (Section~\ref{sec:critic}). Because no candidates have been fitted in the first round, all configurations initially receive the same problem-only prompt. The complete prompt templates are provided in Appendix~\ref{app:prompts}.

Each iteration generates $K=8$ candidate equation structures with free coefficients named $c_0,c_1,\ldots$. This convention distinguishes fitted coefficients from input variables and enables automatic validation of proposed expressions. For example, if the proposer writes $z_3$ instead of the declared input $x_3$, the expression is rejected rather than treating $z_3$ as a new coefficient. Any variable identifier that is neither a declared input nor a valid coefficient name $c_j$ is considered invalid.

To preserve exploration beyond the critic's recommendations, its directives and edits may occupy at most six of the eight proposal slots. The remaining two slots are reserved for proposals that are not conditioned on critic guidance.

\subsection{Stage B: fitting, tempering and feedback labels}
\paragraph{Fitting.} Coefficients are fitted on $\Dfit$ by nonlinear least squares with a trust-region reflective optimiser, started from $6$ initial guesses with the best result kept; each start is limited to $2{,}000$ function evaluations, and no new start begins after $120$\,s on one candidate. The noise estimate of Equation~\ref{eq:residual_variance} is floored at $10^{-12}\,\mathrm{Var}(y)$, where $\mathrm{Var}(y)$ is the variance of the target on $\Dfit$, so that a structure reproducing the data almost exactly does not send $\log\hat\sigma_k^2$ to $-\infty$. The approximation in Equation~\ref{eq:bic} drops terms that are the same for every candidate.

\paragraph{Coefficient uncertainty.} We approximate the uncertainty in the fitted coefficients by a Gaussian centred at $\hat\theta_k$ (a Laplace approximation), whose covariance $\Sigma_k$ is the inverse curvature of the fitting loss at $\hat\theta_k$. $\Sigma_k$ is used only for the feedback labels below, not for the score.

\paragraph{Which coefficients the data determine.}
Two labels, computed from $\Sigma_k$, feed the feedback of Stages A and C. Each coefficient $\theta$ is given a broad reference scale $\max(10, 2|\hat\theta|)$. A coefficient is \emph{unidentified} when its standard deviation under the Laplace approximation is at least half that scale---the data have barely narrowed it, and \emph{insignificant} when it is identified but its fitted value $|\hat\theta|$ is smaller than its standard deviation, so it cannot be told apart from zero. The two call for different repairs: an unidentified coefficient usually signals a redundant term, an insignificant one a term that may not be needed.

\paragraph{Tempering.} At each round $\tau$ is found by bisection so that $B_{\mathrm{eff}}$ of Equation~\ref{eq:effective_models} equals $0.35B$. The resulting temperatures are large, with medians between $700$ and $1{,}900$ across held-out configurations, reflecting BIC differences of thousands between retained structures.

\paragraph{Structural spread.} The spread of Equation~\ref{eq:structural_spread} is computed as $\sigma_S^2(x)=\sum_k w_k(M_k(x;\hat\theta_k)-\hat y(x))^2+10^{-6}\sum_k w_k\hat\sigma_k^2$. The second term only keeps $\sigma_S$ strictly positive where all structures agree. Multiplying $\sigma_S$ by a constant has no effect, because the constant cancels in $\hat q_\alpha$; only its variation across inputs matters, which is why no sizeable constant noise term is included.

\subsection{Stage C: the critic}
\paragraph{The diagnostics are measured from the data, not judged by a model.}
Language models reason poorly over tables of numbers and well over named patterns, so the analysis is computed numerically and the critic is given conclusions. For each candidate in the buffer we report:
\begin{enumerate}[leftmargin=1.6em,itemsep=1pt,topsep=2pt]
\item systematic residual trends against each input, reported when the Spearman rank correlation between residual and input is at least $0.2$ in magnitude, or when mean residuals over five quantile bins span at least $0.5$ residual standard deviations without being monotone;
\item which coefficients are unidentified or insignificant (Stage B);
\item how often the protected operators had to clamp their argument, reported when this affects at least $5\%$ of rows;
\item (rung A3 only) how far the three highest-weight structures diverge from one another on $\Dcext$: for each pair, the NMSE between their two predictions there, beside the same quantity on the training inputs.
\end{enumerate}
At the buffer level we report, in rung A3, the median half-width of the prediction intervals of Stage D on $\Dcext$, relative to the training target's standard deviation, and how much the structural spread $\sigma_S$ (the disagreement among retained structures, Equation~\ref{eq:structural_spread}) grows between the two; and, in all critic rungs, a \emph{stagnation flag}. The whole summary is written in JavaScript Object Notation (JSON), a structured text format, and capped at $4{,}500$ characters, about $1{,}500$ tokens (the word pieces a language model reads).

Stagnation is detected numerically, the best score has not strictly improved for $10$ rounds, and stated to the critic, which chooses between an \emph{explore} mode (ask the proposer for structures unlike the current ones) and an \emph{exploit} mode (refine the current leaders). The mode is \emph{not} forced. On the development set the critic chose explore in about $1$ call in $24$, and Appendix~\ref{app:example} shows a $62$-round plateau on which the flag never fired at a critic call. We report this as a limitation (\S\ref{sec:conclusion}).

\paragraph{The critic's objective.}
The objective is stated in the critic's system prompt rather than computed. In rung A3 it is told that the goal is structures that extrapolate with narrow, honest intervals; to prefer edits and directives that would make the leading structures agree beyond the data; and never to propose anything whose only effect would be a wider interval. In rung A2 the corresponding rule asks for compact, physically plausible forms that fit the data. With no labels on $\Dcext$, agreement among near-equally-supported structures is the strongest available surrogate for being right there: when structures indistinguishable on the data diverge sharply just outside it, at most one of them can be extrapolating correctly.
Ordering the rule this way---agreement first, narrowness second---matters, because either half on its own is degenerate. Asking only for narrow intervals rewards confident nonsense: a single over-fitted structure has no one to disagree with, so $\sigma_S$ is small and the interval is tight wherever it is wrong. Asking only for agreement rewards a buffer of near-copies, which agree everywhere and say nothing. Requiring agreement among structures the evidence cannot separate, and only then preferring the narrow ones, is the combination neither degenerate strategy satisfies. Note that no \emph{coverage} appears in this rule: $\Dcext$ has no labels, so coverage there cannot be computed, and the attainment-and-sharpness pair of Section~\ref{sec:protocol} is an evaluation criterion, not the critic's objective.
\paragraph{Output, validation and memory.}
The critic runs at sampling temperature $0$ (its most likely reply, so that runs are repeatable) with a fixed seed, and must answer in a fixed JSON format; it receives one language-model call in every five of the problem's $125$-call budget: $24$ critic calls and $101$ proposer rounds. We run it with two models: the proposer itself (Llama-3.1-8B-Instruct on the same local server) and GPT-4o.
Its reply names a mode; up to $5$ structural families the proposer should stop proposing (\emph{bans}), which must be copied from the evidence's list of
repeatedly proposed families; $2$--$3$ \emph{directives}, each a short instruction to the proposer (for example, ``try a saturating term in $t$'') with the number of the $8$ slots it should fill; and up to $3$ \emph{edits}, each a complete rewrite of a named buffer entry.

Every edit and directive must cite a piece of the evidence it was shown, and that piece must actually be present; items that cite something absent are dropped and counted. Edits are also checked against the action the critic says it is taking. An edit that claims to \emph{delete a term} must remove one additive term (two from a structure of four or more), or remove the part of a single term holding a coefficient the evidence lists as weak; one that claims to \emph{merge coefficients} must keep the same inputs and use fewer coefficients; one that claims to \emph{replace a term} must swap out at most one term. Valid edits take slots in the next round's batch of $8$ and are fitted, never inserted.

The critic's guidance holds until its next call; a failed call spends its budget and leaves no guidance. There is no persistent archive: the posterior buffer, which the critic may read but never write, is the only memory. Prompt and schema are in Appendix~\ref{app:prompts}.

\subsection{Stage D: additional notes}
\label{app:stageD}
\textbf{A shift-aware variant that did not help.}
Weighted conformal intervals \citep{tibshirani2019conformal}, which reweight the calibration residuals by a logistic-regression density ratio toward the
unlabelled shifted inputs, were evaluated offline on the development buffers. They \emph{lowered} out-of-distribution coverage (A0: $0.61 \to 0.36$--$0.44$), so they are not carried forward.

\textbf{Accuracy and calibration under extrapolation.}
Recovering the correct equation structure can reduce extrapolation error, but it does not guarantee calibrated uncertainty. Even with the correct functional form, residual distributions may differ between calibration and shifted test data because of input-dependent noise or coefficient-estimation errors. Consequently, accurate point predictions do not necessarily yield reliable prediction intervals under extrapolation. LLM-SR-UQ addresses these objectives separately: candidate structures are retained based on fitting evidence and complexity, while prediction intervals are calibrated using held-out observations and scaled by structural disagreement. This approach does not guarantee OOD coverage, which is evaluated empirically. When multiple structures fit the observations similarly, their disagreement provides additional information about uncertainty beyond the observed range.

\begin{figure*}[h]
\centering
\includegraphics[width=0.8\textwidth]{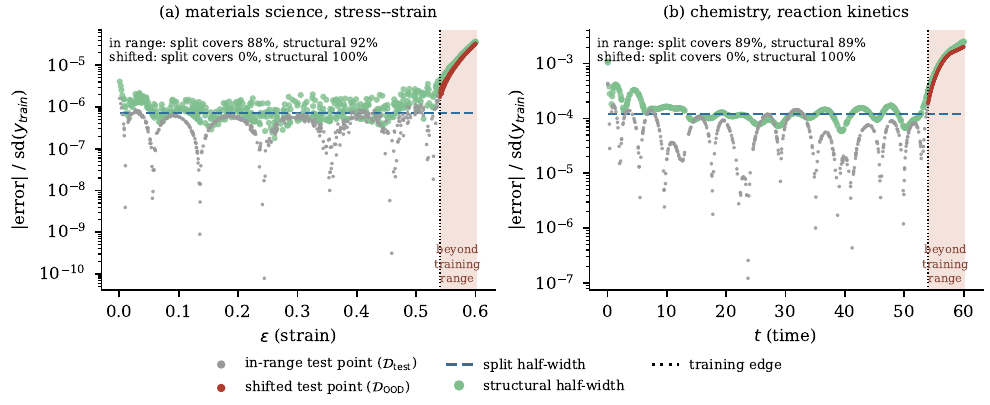}
\caption{Two held-out problems at A1, error against the most-shifted input.}
\label{fig:example}
\end{figure*}

\textbf{Illustrative prediction intervals under extrapolation.} Figure~\ref{fig:example} illustrates how the two interval constructions behave on held-out problems from materials science and chemistry. Within the training range, prediction errors are small and both intervals remain narrow. Under extrapolation, the fixed-width split-conformal intervals fail to cover all shifted points in both examples. Structural intervals widen as the retained equations disagree and cover all shifted points in these cases. Because both methods use identical point predictions, these examples illustrate the effect of adapting interval widths to structural disagreement rather than improving predictive accuracy.

\subsection{The loop in full}

\begin{algorithm2e}[H]
\caption{LLM-SR-UQ. All fitting, scoring and calibration are numerical
computations on the data; the two language models see only text.}
\label{alg:main}
\SetAlgoLined
\KwIn{variable descriptions; splits $\Dfit,\Dcid,\Dcext$; iterations $T$;
      batch $K{=}8$; level $1{-}\alpha$}
\KwOut{weighted buffer $\{(M_k,\hat\theta_k,w_k)\}$ and calibrated intervals}
buffer $\gets \emptyset$; \quad guidance $\gets \emptyset$\;
\For{$t \gets 1$ \KwTo $T$}{
  \textbf{A.} prompt $\gets$ context ($+$ A1 feedback block) ($+$ latest critic
      guidance) \tcp*{context only at $t{=}1$}
  \hphantom{\textbf{A.}} $\{M_1,\dots,M_K\} \sim \text{Proposer}(\text{prompt})$\;
  \textbf{B.} \ForEach{$M_k$}{
      fit $\hat\theta_k,\Sigma_k$ on $\Dfit$; compute $\log Z_k$
      \eqref{eq:bic}; $\ell_k \gets \log Z_k + \log p(M_k)$\;
      insert into buffer if structurally distinct and high-scoring\;
  }
  \hphantom{\textbf{B.}} solve for $\tau$ s.t. $B_{\mathrm{eff}}$ hits target
      \eqref{eq:effective_models}; update $w_k$ \eqref{eq:tempered_weights}\;
  \textbf{C.} \If{this call is a critic call (every 5th of the budget, never the last)}{
      diagnostics $\gets$ \textsf{Analyse}(buffer, residuals, $\Sigma$, stagnation)\;
      \hphantom{xx} ($+$ in A3, no labels: top-3 divergence on $\Dcext$,
        $\sigma_S$ growth, half-widths)\;
      guidance $\gets$ \textsf{Critic}(diagnostics); validate grounding, actions,
        slot limits\;
  }
}
\textbf{D.} $\hat y \gets \sum_k \tilde w_k M_k$ at $\tau{=}1$ \eqref{eq:bma_prediction};
   $\sigma_S$ from tempered $w_k$ \eqref{eq:structural_spread}; calibrate normalized scores
   on $\Dcid$ \eqref{eq:conformal_score}; report coverage and width on $\Dtest$ and
   $\Dood$\;
\end{algorithm2e}

\noindent All rungs have the same budget of $125$ language-model calls per
problem. In rungs A2 and A3 one call in five goes to the critic, so they run
$101$ proposer rounds ($808$ candidates) against $125$ rounds ($1{,}000$
candidates) for A0 and A1. A worked example of one run is given in
Appendix~\ref{app:example}.

\section{The retained buffer: concentration and recovery}\label{app:posterior}
 
This appendix supports two claims made in Section\ref{sec:bayes} and Section\ref{sec:calib-results}: that the evidence weights are effectively degenerate at this sample size, and that the calibration gain does not depend on their being otherwise. All numbers come from the exported buffers of the six held-out configurations; no search was re-run and no language-model call was made.
 
\paragraph{Concentration.}
Untempered weights are $\tilde w_k \propto \exp(\ell_k)$, recomputed from the stored BIC score with the per-coefficient penalty re-applied. Effective family count is $(\sum_k \tilde w_k^2)^{-1}$.
 
\begin{table}[h]
\centering\small
\caption{Concentration of the retained buffer. The tempered columns are pinned by
construction ($B_{\mathrm{eff}} = 8.4$) and are shown only so the gap is visible;
the untempered columns are what the evidence says.}
\label{tab:concentration}
\begin{tabular}{lccccc}
\toprule
& & \multicolumn{2}{c}{untempered} & \multicolumn{2}{c}{tempered} \\
\cmidrule(lr){3-4}\cmidrule(lr){5-6}
Configuration & $n$ & median $\tilde w_1$ & median $B_{\mathrm{eff}}$ & median $w_1$ & median $B_{\mathrm{eff}}$ \\
\midrule
A0                   & 109 & 1.0000 & 1.00 & 0.240 & 8.40 \\
A1                   & 109 & 1.0000 & 1.00 & 0.227 & 8.40 \\
A2 ($8$B critic)     & 108 & 1.0000 & 1.00 & 0.229 & 8.40 \\
A3 ($8$B critic)     & 107 & 1.0000 & 1.00 & 0.219 & 8.40 \\
A2 (GPT-4o critic)   & 107 & 1.0000 & 1.00 & 0.243 & 8.40 \\
A3 (GPT-4o critic)   & 108 & 1.0000 & 1.00 & 0.231 & 8.40 \\
\bottomrule
\end{tabular}
\end{table}
 
\noindent The untempered top-family weight exceeds $0.95$ on $468$ of $647$ problem-configurations. The solved temperatures have medians between $700$ and $1{,}900$, which is the scale required to flatten log-evidence gaps of thousands of nats. Concentration is lowest in materials science and chemistry, where the top weight falls below $0.5$ on roughly a third of problems, and highest in physics ($35$ of $39$ above $0.95$ at A0).
 
\paragraph{The gain does not require spread.}
Splitting the held-out problems by whether the untempered buffer is degenerate ($B_{\mathrm{eff}} \le 1.01$, i.e.\ the evidence picks exactly one structure):
 
\begin{table}[h]
\centering\small
\caption{Attainment under split and structural intervals, split by whether the evidence weights are degenerate. The construction delivers its gain in both regimes.}
\label{tab:gain-by-conc}
\begin{tabular}{llccc}
\toprule
Rung & Subset & $n$ & $A_{0.1}$ split & $A_{0.1}$ $S$ \\
\midrule
A0 & evidence degenerate & 86 & 0.58 & 0.86 \\
A0 & some spread         & 23 & 0.57 & 0.74 \\
A1 & evidence degenerate & 62 & 0.57 & 0.88 \\
A1 & some spread         & 47 & 0.49 & 0.98 \\
\bottomrule
\end{tabular}
\end{table}
 
\noindent The Spearman correlation between untempered $B_{\mathrm{eff}}$ and the
per-problem coverage gain is $-0.12$ ($p=0.23$) at A0 and $+0.15$ ($p=0.13$) at
A1. Whatever $\sigma_S$ is measuring, it is not posterior spread.
 
\paragraph{Recovery.}
Functional recovery at $k$, over problems whose ground-truth structure can
reproduce its own data to $\mathrm{NMSE}<10^{-8}$ (four chemistry problems fail
this in every configuration and are excluded).
 
\begin{table}[h]
\centering\small
\caption{Functional recovery by rank cutoff. The column that matters is the
last: problems on which the retained set contains a recovery that its
top-ranked member misses.}
\label{tab:recovery-k}
\begin{tabular}{lccccc}
\toprule
Configuration & with ref. & FR@1 & FR@3 & FR@10 & FR@3 $\wedge\,\neg$FR@1 \\
\midrule
A0                    & 105 &  8 &  8 &  8 & 0 \\
A1                    & 104 & 12 & 12 & 12 & 0 \\
A2 ($8$B critic)      & 104 & 13 & 14 & 15 & 1 \\
A3 ($8$B critic)      & 103 & 11 & 13 & 13 & 2 \\
A2 (GPT-4o critic)    & 103 & 10 & 10 & 10 & 0 \\
A3 (GPT-4o critic)    & 104 & 10 & 11 & 12 & 1 \\
\bottomrule
\end{tabular}
\end{table}
 
\noindent No McNemar test on FR@3 against FR@1 is significant (exact $p=1.0$,
$0.5$, $1.0$ where a discordant pair exists).
 
\paragraph{Sensitivity to the tempering target.}
Since the spread is produced by tempering rather than by the evidence, the target
$B_{\mathrm{eff}} = 0.35B$ is the parameter the construction rests on, and it was
fixed on the development set. We therefore re-weighted every exported held-out
buffer at nine targets from $B_{\mathrm{eff}} = 1$ (a point mass) to
$B_{\mathrm{eff}} = 24$ (uniform over a full buffer), re-solving $\tau$,
recomputing $\sigma_S$, recalibrating on $\Dcid$ and re-evaluating, with the
structures, fits, splits and seeds untouched. Because the predictive mean uses
the untempered weights, the point predictions are identical at every target and
the split-conformal reference is a single number per configuration: the sweep
moves the interval scale and nothing else. At $B_{\mathrm{eff}} = 8.4$ the
procedure reproduces the published attainment to a mean absolute per-problem
coverage difference of $0.0003$ or better.
 
\begin{figure}[h]
\centering
\includegraphics[width=0.78\linewidth]{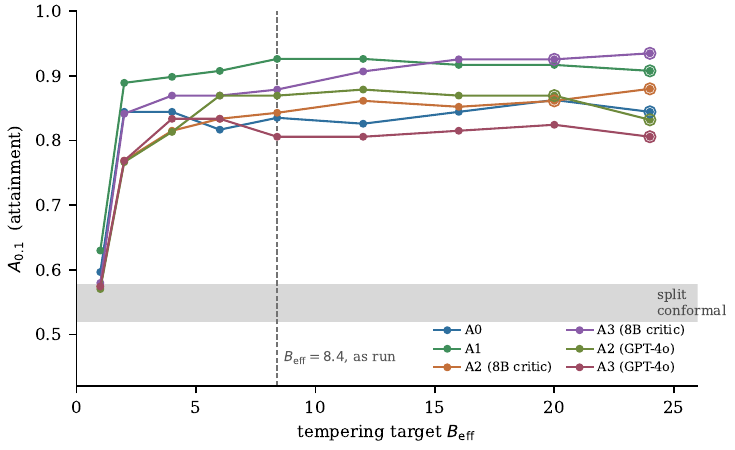}
\caption{Attainment against the tempering target, held-out, all six
configurations. The grey band spans the split-conformal reference across
configurations. The entire effect is the step from a point mass
($B_{\mathrm{eff}}=1$) to any genuine spread; above $B_{\mathrm{eff}}=2$ the
curves are flat to within $0.11$ and every one of them beats split conformal by
$20$--$40$ points. Hollow rings mark targets the bisection could not reach for
some problems, which is why the $B_{\mathrm{eff}}=24$ column is not a clean
uniform-weighting condition: it is unreachable on $9$--$46$ problems per
configuration, against at most $5$ at every smaller target.}
\label{fig:beff}
\end{figure}
 
\noindent The curve is flat, not peaked, and the value we ran is not the optimum
--- several larger targets do slightly better on attainment and on sharpness.
This is worth stating plainly in both directions. It means $B_{\mathrm{eff}}$ is
not a tuned hyperparameter in the sense that would compromise the result: a
quantity tuned on the development set would show a peak at or near the chosen
value, and this one does not. It also means we cannot claim $0.35B$ is
principled. The defensible claim is the measured one: \emph{attainment is
insensitive to the tempering target over an order of magnitude, provided the
weights are not collapsed onto a single structure.}
 
The degenerate subset makes the point most sharply. On the $86$ A0 problems where
the evidence is a point mass and tempering therefore supplies all of the spread,
attainment is $50/86$ at $B_{\mathrm{eff}}=1$ --- indistinguishable from split
conformal's $50/86$, as it should be --- and $73$ to $75$ of $86$ at every target
from $2$ to $24$, with no dependence on where in that range the target sits. The
mechanism needs a set with spread. It does not need a particular amount of it.
 
\paragraph{Two measurement bounds.}
First, per-member evidence within a structural family is not exported, only the
family's best member. A family's rank aggregates its members' summed tempered
weight, so the untempered weight computed here is a lower bound for
multi-member families; some family has more than one member on $160$ of $647$
problem-configurations, and the exported rank $1$ differs from the
best-member-evidence argmax on $16$. Second, distinct structural families are not
necessarily distinct functions. On one held-out biology problem whose top family
holds only $0.14$ of the untempered weight --- which would read as a genuinely
undecided posterior --- the five highest-weighted families carry \emph{identical}
weights and agree with one another to $\approx 10^{-13}$ at every point of a grid
spanning the training range and well beyond it. They are one function written
five ways. A low top-family weight is therefore not by itself evidence of
functional diversity, and we do not use it as such anywhere in this paper. How
often this occurs across the benchmark is not something the current export can
answer, and we flag it as a measurement limitation rather than estimating it.

\section{Does the calibration-aware critic improve search?}
\label{app:critic-results}

\begin{table*}[t]
\centering\small
\caption{Ablation ladder. Every rung reports coverage, so uncertainty is not a
late addition, and every rung has the same budget of $125$ language-model calls.
A1 is the control: if A2 $\approx$ A1, mechanical feedback was the whole story.}
\label{tab:ladder}
\begin{tabular}{llll}
\toprule
Rung & Feedback to proposer & Critic & Isolates \\
\midrule
A0 & frozen prompt (problem description only) & --- & search + calibration \\
A1 & + mechanical diagnostics (fixed template) & --- & plumbing \\
A2 & + critic guidance from point-fit diagnostics & yes & critic from fit alone \\
A3 & + label-free calibration diagnostics (divergence & yes & \textbf{the claim} \\
   & \quad beyond the data, interval growth) & & \\
\bottomrule
\end{tabular}
\smallskip
\end{table*}

The ablation ladder isolates the critic at a fixed budget of $125$ language-model
calls per problem; in A2 and A3, one call in five goes to the critic
(Table~\ref{tab:ablation}). We ran it twice: with the proposer model as its own
critic, and with GPT-4o as critic. The primary evidence is held-out ($109$
problems, one seed, all six rungs); the development set ($20$ problems, five
seeds) supplies seed replication and the mechanism counters.
 
\begin{table}[t]
\centering\small
\setlength{\tabcolsep}{3.5pt}
\caption{The ablation ladder, held-out, at a matched budget of $125$
language-model calls per problem (A2/A3: $101$ proposer rounds and $24$ critic
calls). NMSE are medians; coverage and $A_{0.1}$ use structural intervals, and
the split column gives $A_{0.1}$ of split conformal around the same predictor.
Parameters are the mean coefficient count of the reported equation. Plateau is
the fraction of rounds after the last improvement of the best score.}
\label{tab:ablation}
\begin{tabular}{lcccccccc}
\toprule
& \multicolumn{2}{c}{NMSE} & OOD & \multicolumn{2}{c}{$A_{0.1}$} & & & \\
\cmidrule(lr){2-3}\cmidrule(lr){5-6}
Rung & ID & OOD & cov. & $S$ & split & $S_{0.1}$ & FR@1 & params / plateau \\
\midrule
A0 & 3.3e-6 & 2.4e-3 & 0.906 & 0.83 \tiny(91/109) & 0.58 & 1.27e-2 &  8/105 & 6.7 / 0.52 \\
A1 & 2.0e-8 & 6.0e-5 & 0.964 & \textbf{0.92} \tiny(100/109) & 0.53 & 1.12e-3 & 12/104 & 8.2 / 0.42 \\
\midrule
\multicolumn{9}{l}{\emph{critic $=$ the proposer model (Llama-3.1-8B-Instruct)}}\\
A2 & 9.3e-9 & 2.0e-5 & 0.924 & 0.87 \tiny(91/105) & 0.52 & 1.87e-3 & 13/104 & 8.1 / 0.46 \\
A3 & 4.4e-7 & 4.0e-4 & 0.947 & 0.88 \tiny(93/106) & 0.56 & 7.73e-3 & 11/103 & 7.4 / 0.48 \\
\multicolumn{9}{l}{\emph{critic $=$ GPT-4o}}\\
A2 & 5.7e-7 & 1.5e-4 & 0.938 & 0.87 \tiny(93/107) & 0.52 & 3.43e-3 & 10/103 & 6.9 / 0.52 \\
A3 & 1.1e-6 & 1.3e-4 & 0.927 & 0.82 \tiny(88/107) & 0.56 & 2.39e-3 & 10/104 & 6.7 / 0.43 \\
\bottomrule
\end{tabular}
\end{table}
 
\begin{figure}[t]
\centering
\includegraphics[width=\linewidth]{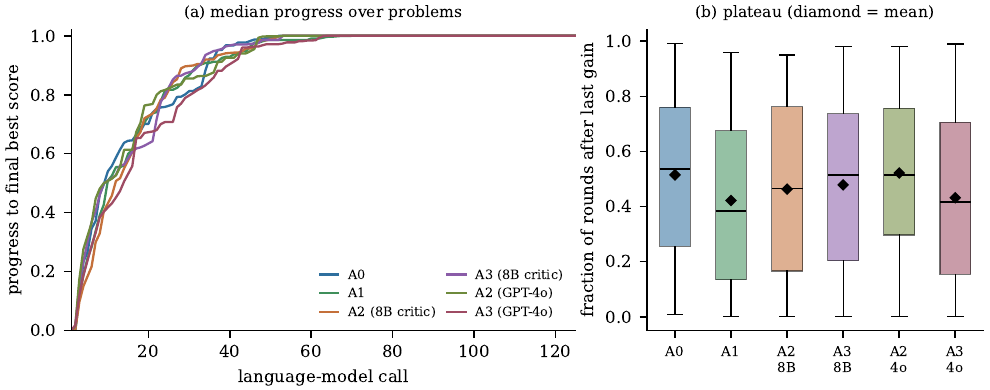}
\caption{Search behaviour across the ladder, held-out. \textbf{(a)} Progress of the best buffer score, normalized per problem so that $0$ is the first round and $1$ the final best, plotted against the language-model call index (critic calls included), median over problems. \textbf{(b)} Fraction of proposer rounds spent after the last improvement, per problem. The six configurations are nearly indistinguishable: every one reaches $90\%$ of its final score by a median of $32$ to $41$ calls and then spends roughly half its budget without improving.}
\label{fig:plateau}
\end{figure}

\paragraph{Mechanical feedback helps.}
Search dynamics are almost the same on every rung (Figure~\ref{fig:plateau}):
each reaches $90\%$ of its final score by a median of $32$ to $41$ calls and then
spends roughly half its budget without improving, so differences between rungs
come from \emph{what} is found, not from how fast. A1 against A0: out-of-distribution NMSE improves ($p=7\times10^{-9}$, paired) and attainment rises from $0.83$ to $0.92$. Functional recovery moves from $8$ to $12$ of $104$, which is not significant (McNemar $p=0.29$), and the equations get somewhat larger ($6.7 \to 8.2$ coefficients, $p=2\times10^{-7}$).
 
\paragraph{The critic does not help, with either model.}
No critic rung matches A1. Attainment is $0.87$ and $0.88$ with the $8$B critic and $0.87$ and $0.82$ with GPT-4o, against A1's $0.92$. Against A1, the only significant coverage changes go the wrong way: $0.963 \to 0.923$ at A2 with the $8$B critic ($p=0.019$) and $0.964 \to 0.930$ at A3 with GPT-4o ($p=0.030$), both still above nominal. Functional recovery, width and NMSE are otherwise flat at A2. These results suggest that mechanical feedback provides the main benefit of search feedback under the tested computational budget.
 
\paragraph{Conditioning the critic on calibration evidence makes intervals wider,
not sharper.} With the $8$B critic, A3 against A2 widens the shifted intervals ($p=0.018$) and worsens NMSE in and out of distribution ($p=0.0068$ and $p=0.0061$), with coverage and recovery unchanged. With GPT-4o, A3 changes nothing beyond A2 (all $p>0.3$). Thus, additional calibration-aware critic feedback does not improve these outcomes and, with the 8B critic, worsens interval width and prediction accuracy. The development set agrees ($p=0.058$ on width, $p=0.009$ on fit), so this is not a single-seed artifact.
 
\paragraph{What the critic does change is simplicity, and it is paid for.} Both critics return significantly simpler equations than mechanical feedback alone -- $7.4$ coefficients with the $8$B critic ($p=0.0055$) and $6.7$ with GPT-4o at A3 ($p=1.4\times10^{-9}$), against A1's $8.3$ --- and both lose in-distribution accuracy in doing so ($p=6.5\times10^{-5}$ and $p=2.5\times10^{-4}$).
 
\paragraph{The null is not a critic that never fired.} On the development set the
critic makes $24$ calls and proposes about $70$ edits per problem-run. GPT-4o's edits pass the mechanical checks $78.5\%$ of the time against $29.1\%$ for the $8$B model, enter the retained buffer about twice as often ($2.4$ against $1.4$ per run at A2, $3.1$ against $1.1$ at A3), cite missing evidence on $0.3\%$ of items against $8.4\%$, and choose exploration in most calls rather than almost none. The critic works, and works better with a stronger model. The outcome metrics still do not move, and the buffer explains why: an edit must beat the incumbent structures on evidence to be kept, and only one to three of about twenty valid edits per run do. A critic that proposes well but rarely displaces anything cannot change what the buffer determines.
 
\paragraph{Reading the null.}
Two of the three hypotheses this paper set out to test were not supported, and it is worth saying plainly what that leaves standing. What fails is the claim that a language model, shown diagnostics about its own search, will steer that search toward better-calibrated equations. It does not, for either critic model, on either data set, under any metric we fixed in advance, and the design has the power to show it: $109$ paired held-out problems, five development seeds, two critic models an order of magnitude apart in capability, and counters confirming the critic was active. What stands is independent of the critic. The mismatch between accuracy and calibration under shift is a property of the benchmark and the methods evaluated on it, the interval construction is what delivers calibration (Section\ref{sec:calib-results}); and it works on searches we did not write (Section\ref{sec:transfer}). A reader who discards the critic entirely still keeps the measurement, the construction and the wrapper.

\section{Prompts and the critic contract}
\label{app:prompts}
 
Every prompt and reply in this appendix is reported exactly as the models sent and received it. Where a value was not recorded, we say so rather than reconstructing it.
 
 
\subsection{Proposer system message}
 
The proposer is given these instructions before every round. The one placeholder, the number of expressions requested, is the batch of $8$ less any critic edits being fitted that round.
 
\begin{Verbatim}[fontsize=\footnotesize,breaklines=true,frame=single,framesep=3pt]
You propose candidate symbolic-regression expressions for a scientific
equation-discovery task.
Return JSON with a single key 'expressions': a list of exactly {n} strings.
 
Variables - use ONLY these, spelled exactly: {var_names}.
Coefficients - every fitted constant must be named c0, c1, c2, ...
Numbering restarts at c0 in each expression.
Any identifier that is neither a listed variable nor a c<number>
coefficient is rejected, so do not invent symbol names.
 
Allowed functions: sin cos tan tanh sinh cosh exp log sqrt abs sign
softplus minimum maximum.
Write exponentiation as ** (Python syntax), never ^. Give the
right-hand side only, with no 'y =' prefix.
Use softplus(...) where a quantity must stay positive, such as the base
of a fractional power or the argument of a logarithm.
 
Example of the required form: c0*exp(-c1*t) + c2
Return JSON only, with no prose and no comments.
\end{Verbatim}
 
\noindent Responses are constrained by a JSON schema requiring exactly $n$ strings under the key \texttt{expressions}. Both backbones are given the same schema so neither has an easier decoding task.
 
\subsection{A complete rendered prompt}
\label{app:rendered}
 
Round $49$ of $101$ on a population-growth problem, in the calibration-aware configuration. This round directly follows a critic call, so the prompt shows all three of its parts at once: the problem description, the mechanical feedback block, and the critic's guidance.
 
\begin{Verbatim}[fontsize=\footnotesize,breaklines=true,frame=single,framesep=3pt]
Discover the mathematical equation relating the following variables:
Output: dP_dt - Population growth rate
Input 1: t - Time
Input 2: P - Population at time t
 
Propose candidate closed-form expressions for dP_dt in terms of t, P.
Prefer forms that are physically plausible for this domain, and keep them
numerically well behaved across the whole input range.
 
Feedback on the candidates fitted so far (384 fitted over 48 rounds):
 
Best structures so far (NMSE on the training data; lower is better):
  1. c0+c1*P**c2+c3*P**c4+c5*c6**c7/(c8+t**c9)   NMSE=9.5e-09  coefficients=10
  2. c0+c1*P**c2+c3*t**c4+c5*P**c6   NMSE=2.39e-06  coefficients=7
  3. c0+c1*P**c2+c2*P**c3/(c4+t**c5)   NMSE=3.07e-06  coefficients=6
 
Coefficients the data does not determine; the term they sit in is doing no
work, so remove or rewrite it:
  - structure 1: c5 in `c5 * c6 ** c7 / (c8 + t ** c9)` (not constrained by the data)
  - structure 1: c6 in `c5 * c6 ** c7 / (c8 + t ** c9)` (not constrained by the data)
  - structure 1: c7 in `c5 * c6 ** c7 / (c8 + t ** c9)` (not constrained by the data)
 
Residual of structure 1 (residual scale 9.6e-05 of the target's):
  - falls with t (rank correlation -0.22): the dependence on t is missing or misshaped
  - falls with P (rank correlation -0.23): the dependence on P is missing or misshaped
 
Tried repeatedly without reaching the best structures; do not propose these
families again:
  - c0+c1*t**c2+c3*P**c4   (24 times)
  - c0+c1*P**c2+c3*t**c4   (18 times)
  - c0+c1*P**c2+c3*P**c4   (13 times)
  - c0+c1*(P**c2)/(c3+t**c4)   (10 times)
  - c0+c1*P**c2   (9 times)
 
Critic guidance:
The search should exploit: refine the best structures rather than start over.
These rewrites are already being fitted this round; do not repeat them:
  - c0+c1*P**c2+c3*P**c4
  - c0+c1*P**c2+c3*t**c4
Write your 6 expressions in this order:
  - 2 expressions: propose forms with fewer coefficients
  - 2 expressions: propose forms with physically motivated terms
  - the remaining 2: your own choice.
Do not propose these families again:
  - c0+c1*t**c2+c3*P**c4
  - c0+c1*P**c2+c3*t**c4
  - c0+c1*P**c2+c3*P**c4
  - c0+c1*(P**c2)/(c3+t**c4)
  - c0+c1*P**c2
\end{Verbatim}
 
\noindent The model's reply to this prompt, verbatim:
 
\begin{Verbatim}[fontsize=\footnotesize,breaklines=true,frame=single,framesep=3pt]
{"expressions": ["c0+c1*P**c2", "c0+c1*t**c2+c3*P",
 "c0+c1*P**c2/(c3+t**c4)", "c0+c1*P**c2+c2*P**c3/(c4+t**c5)",
 "c0+c1*exp(-c2*t)*P**c3", "c0+c1*P**c2+c3*tanh(c4*t)*P**c5"]}
\end{Verbatim}
 
\noindent All six parse and fit. The prompt's structure is visible in the reply: the first two expressions are short, answering the ``fewer coefficients`` directive, and the later ones reuse the saturating and exponential forms already in the buffer.
 
\subsection{Critic input, output, and validation}
 
\paragraph{The critic's instructions.}
The critic receives the following system message before every call. Placeholders are filled from the problem: the variable list, the batch size, and---in the calibration-aware configuration only---the final rule, which is omitted otherwise. The text below is the version used in the runs we report, confirmed against the recorded hash of the code that produced them.
 
\begin{Verbatim}[fontsize=\footnotesize,breaklines=true,frame=single,framesep=3pt]
You are the critic in an equation-discovery loop. A proposer writes candidate
expressions; each is fitted to the data and scored by Bayesian evidence, and
only the evidence decides what is kept. You never decide what is kept. You read
the evidence JSON and write edits and directives for the proposer.
 
Rules.
1. Ground everything. Every edit and directive names an evidence_field: a path
into the evidence JSON, with dots for keys and [n] for list positions, such as
weak_coefficients[0], residual.patterns[1], variables.inclusion.t or stagnation.
An item whose evidence_field does not exist, or points to an empty value, is
discarded.
2. Edits (at most 3) are complete rewrites of one of top_structures, identified
by of_rank, and are fitted like any proposal. The expression is always the WHOLE
structure after the change - never a single term, and never the structure
unchanged. Actions:
   delete_term: remove one term whose coefficient the data does not determine
(two, if the structure has four or more terms), or remove the part of one term
that holds such a coefficient (c0*P/(c1 + c2*t) becomes c0*P when
weak_coefficients names c1); keep everything else, and keep the names of the
coefficients you keep.
   merge_coefficients: rewrite with fewer coefficients where several only act
together (c0*c1*P becomes c0*P); keep the same inputs.
   replace_term: swap exactly one term the residual shows is misshaped for a
better one; keep the other terms.
   Example: top structure 1 is c0*P + c1*P**2 + c2*sin(c3*t) and
weak_coefficients[0] names c2 in c2*sin(c3*t). A valid edit is {"of_rank": 1,
"action": "delete_term", "expression": "c0*P + c1*P**2", "evidence_field":
"weak_coefficients[0]"}.
   An edit that does not match its action is discarded.
3. Directives (2 or 3) steer the proposer's next rounds. Make them pursue
different hypotheses. Each takes n_slots of the batch of 8; directive slots plus
edits may not exceed 6, so at least 2 proposals stay unconditioned.
4. banned_families (at most 5): expressions copied from repeated_families that
the proposer should stop proposing. Nothing else may be banned.
5. mode: "exploit" to refine the current best structures; "explore" when the
search has stagnated or the best structures look physically implausible.
6. The evidence has a calibration block, computed without any held-out labels:
how far the top structures' predictions diverge beyond the training inputs, and
how much the prediction interval widens there. The goal is structures that
extrapolate with narrow, honest intervals. Prefer edits and directives that
would make the top structures agree beyond the data - forms that discriminate
between them, or physically motivated forms - and never propose anything whose
only effect would be a wider interval.
 
Expressions: use only the variables t, P, and coefficients named c0, c1, c2, ...
Allowed functions: sin cos tan tanh sinh cosh exp log sqrt abs sign softplus
minimum maximum. Write powers as **. Right-hand side only.
Return JSON only.
\end{Verbatim}
 
\noindent Rule 6 is the only difference between the two critic configurations. Without calibration evidence it reads instead: \emph{``The goal is a law, not a curve fit: compact, physically plausible forms that fit the data.''} Everything else is identical, so the two configurations differ in exactly one paragraph of instruction and in the presence of one block of evidence.
 
The user message is the problem description followed by the evidence, serialised as JSON. It is not stored as a string, but it is reconstructible from the stored evidence and the problem description; rebuilding it reproduces the recorded character count for all $6{,}696$ critic calls in these runs, with no mismatches.
 
\paragraph{What the critic received.}
The evidence the critic was given before round $49$, exactly as sent. Numbers are rounded to four significant figures.
 
\begin{Verbatim}[fontsize=\scriptsize,breaklines=true,frame=single,framesep=3pt]
{"iteration": 48, "candidates_fitted": 384,
 "last_batch": {"proposed": 8, "fitted": 8, "entered_buffer": 0, "rejections": []},
 "top_structures": [
   {"rank": 1, "expression": "c0+c1*P**c2+c3*P**c4+c5*c6**c7/(c8+t**c9)",
    "train_mse": 6.665e-08, "train_nmse": 9.497e-09, "n_params": 10, "weight": 0.3118},
   {"rank": 2, "expression": "c0+c1*P**c2+c3*t**c4+c5*P**c6",
    "train_mse": 1.678e-05, "train_nmse": 2.391e-06, "n_params": 7, "weight": 0.0679},
   {"rank": 3, "expression": "c0+c1*P**c2+c2*P**c3/(c4+t**c5)",
    "train_mse": 2.155e-05, "train_nmse": 3.071e-06, "n_params": 6, "weight": 0.06543}],
 "weak_coefficients": [
   {"rank": 1, "coefficient": "c5", "kind": "unidentified",
    "terms": ["c5 * c6 ** c7 / (c8 + t ** c9)"]},
   {"rank": 1, "coefficient": "c6", "kind": "unidentified",
    "terms": ["c5 * c6 ** c7 / (c8 + t ** c9)"]},
   {"rank": 1, "coefficient": "c7", "kind": "unidentified",
    "terms": ["c5 * c6 ** c7 / (c8 + t ** c9)"]}],
 "variables": {"inclusion": {"t": 0.8667, "P": 1.0}, "unused_in_top": []},
 "residual": {"expression": "c0+c1*P**c2+c3*P**c4+c5*c6**c7/(c8+t**c9)",
   "residual_rel_sd": 9.618e-05,
   "patterns": [{"variable": "t", "shape": "falls", "rank_corr": -0.2171,
                 "bin_range_sd": 0.2188},
                {"variable": "P", "shape": "falls", "rank_corr": -0.2252,
                 "bin_range_sd": 0.2187}]},
 "protection": [],
 "repeated_families": [{"expression": "c0+c1*t**c2+c3*P**c4", "times": 24},
   {"expression": "c0+c1*P**c2+c3*t**c4", "times": 18},
   {"expression": "c0+c1*P**c2+c3*P**c4", "times": 13},
   {"expression": "c0+c1*(P**c2)/(c3+t**c4)", "times": 10},
   {"expression": "c0+c1*P**c2", "times": 9}],
 "stagnation": {"iterations_since_improvement": 9, "stagnant": false},
 "calibration": {
   "probe": "inputs up to 0.25 training-range widths beyond the data; no labels used",
   "disagreement": [{"ranks": [1, 2], "train_nmse": 2.182e-06, "probe_nmse": 5.981e-05},
                    {"ranks": [1, 3], "train_nmse": 2.714e-06, "probe_nmse": 0.3279},
                    {"ranks": [2, 3], "train_nmse": 4.629e-08, "probe_nmse": 0.3265}],
   "sigma_growth": 750.3,
   "interval_halfwidth_rel": {"calibration": 4.655e-05, "probe": 0.03493}}}
\end{Verbatim}
 
\noindent The final \texttt{calibration} block is what separates the two critic configurations; it is absent without calibration evidence. It is label-free: \texttt{disagreement} compares the top structures against each other on a probe box extending $0.25$ training-range widths beyond
the data, and \texttt{interval\_halfwidth\_rel} is the median conformal half-width relative to the training target's standard deviation. Here the structures agree to $\mathrm{NMSE}\approx 3\times10^{-6}$ on the training data and disagree by $0.33$ on the probe: five orders of magnitude apart.
 
\paragraph{The critic's response.}
\begin{Verbatim}[fontsize=\footnotesize,breaklines=true,frame=single,framesep=3pt]
{"mode": "exploit",
 "edits": [
   {"of_rank": 1, "action": "delete_term", "expression": "c0+c1*P**c2+c3*P**c4",
    "evidence_field": "weak_coefficients[0]"},
   {"of_rank": 2, "action": "merge_coefficients", "expression": "c0+c1*P**c2+c3*t**c4",
    "evidence_field": "residual.patterns[0].rank_corr"},
   {"of_rank": 3, "action": "replace_term", "expression": "c0+c1*P**c2+c3*P**c4",
    "evidence_field": "residual.patterns[1].rank_corr"}],
 "directives": [
   {"instruction": "propose forms with fewer coefficients",
    "evidence_field": "variables.inclusion.t", "n_slots": 2},
   {"instruction": "propose forms with physically motivated terms",
    "evidence_field": "calibration.disagreement[0].train_nmse", "n_slots": 2},
   {"instruction": "propose forms that extrapolate with narrow intervals",
    "evidence_field": "calibration.interval_halfwidth_rel.calibration", "n_slots": 4}],
 "banned_families": ["c0+c1*t**c2+c3*P**c4", "c0+c1*P**c2+c3*t**c4",
   "c0+c1*P**c2+c3*P**c4", "c0+c1*(P**c2)/(c3+t**c4)", "c0+c1*P**c2"]}
\end{Verbatim}
 
\paragraph{Mechanical validation.}
Every response is checked before any of it reaches the proposer. For this call:
 
\begin{Verbatim}[fontsize=\footnotesize,breaklines=true,frame=single,framesep=3pt]
{"parsed": true,
 "dropped": [
   {"kind": "edit", "why": "repeats another edit",
    "item": {"of_rank": 3, "action": "replace_term",
             "expression": "c0+c1*P**c2+c3*P**c4"}},
   {"kind": "directive", "why": "no slots left",
    "item": {"instruction": "propose forms that extrapolate with narrow intervals",
             "n_slots": 4}}],
 "edits_proposed": 3, "edits_valid": 2,
 "directives_proposed": 3, "directives_valid": 2,
 "slots_trimmed": 0, "banned_proposed": 5, "banned_valid": 5, "ungrounded": 0}
\end{Verbatim}
 
\noindent Three checks apply. Duplicate edits are dropped --- here the third repeated the first. Directives are trimmed to the slots actually left: the third asked for four when two remained. And every claim must point at a field that exists in the evidence; on this call all did. Only what survives these checks reaches the proposer, which is why the guidance shown earlier lists two directives rather than the three the critic wrote.
 
\subsection{Aggregate critic behaviour on this problem}
 
\begin{table}[h]
\centering\small
\caption{Critic activity on the problem of Section\ref{app:rendered}: $24$ calls over
$101$ proposer rounds, costing about $49$k tokens in total.}
\label{tab:critic-counters}
\begin{tabular}{lr@{\qquad}lr}
\toprule
critic calls & 24 & edits proposed & 72 \\
failed calls & 0 & edits valid & 44 \\
prompt tokens & 39{,}120 & edits fitted & 44 \\
completion tokens & 10{,}245 & edits entering buffer & 0 \\
directives proposed & 72 & bans proposed & 111 \\
directives valid & 52 & bans valid & 110 \\
slots trimmed & 19 & proposals under ban & 701 \\
ungrounded claims & 2 & ban violations & 80 \\
explore calls & 0 & ungrounded rate & $0.008$ \\
\bottomrule
\end{tabular}
\end{table}
 
These counters exist to check that the critic is doing what it was asked to do rather than producing plausible-looking text: that its claims point at evidence that exists, that its edits and bans are well formed, and that nothing reaches the proposer without passing the checks. Section\ref{sec:protocol} reports them across all problems and configurations.

\section{A run, end to end}
\label{app:example}
 
This appendix follows one problem through the whole loop, using logged values only, to show concretely what each stage produces and how the pieces connect.

\paragraph{Setup.}
A population-growth problem, run in the calibration-aware configuration with one seed: $101$ proposer rounds and $24$ critic calls, $125$ LLM calls in total, with Llama-3.1-8B-Instruct as both proposer and critic. The task is to predict the growth rate $\dot P$ from time $t$ and population $P$. The true equation, which the method never sees, is logistic growth with a sigmoidal harvesting term:
 
\begin{Verbatim}[fontsize=\footnotesize,breaklines=true,frame=single,framesep=3pt]
0.1727856699554779*(1 - P(t)/48.5483027446097)*P(t)
  + 0.1727856699554779*P(t)/(1 + exp(-1.5237423595759925*(-0.9236923548638324 + P(t))))
\end{Verbatim}
 
\paragraph{How the search moved.}
Best log-evidence never decreases, and it improved on only six of the $101$ rounds. Table~\ref{tab:trace} gives every round on which the leading structure changed, plus the last round.
 
\begin{table}[h]
\centering\small
\caption{Every improvement in the leading structure, and the final round.
Buffer size is capped at $24$; the effective model count is pinned at $8.4$ by
the temperature controller of Section\ref{sec:bayes}, as designed.}
\label{tab:trace}
\begin{tabular}{rlrr}
\toprule
round & leading structure & log-evidence & families \\
\midrule
1  & \texttt{c0+c1*(P**c2)/(c3+t**c4)}                       & 5{,}496  & 8 \\
2  & \texttt{c0+c1*P**c2/(c3+t**c4)+c5*P**c6}                & 16{,}847 & 16 \\
5  & \texttt{c0+c1*P**c2+c3*P**c4}                           & 16{,}855 & 24 \\
6  & \texttt{c0+c1*P**c2+c3*P**c4+c5*t**c6}                  & 17{,}207 & 24 \\
7  & \texttt{c0+c1*P**c2+c3*P**c4+c5*softplus(c6*t)**c7}     & 18{,}320 & 24 \\
39 & \texttt{c0+c1*P**c2+c3*P**c4+c5*c6**c7/(c8+t**c9)}      & 26{,}398 & 22 \\
\midrule
101 & \emph{unchanged since round 39}                        & 26{,}398 & 22 \\
\bottomrule
\end{tabular}
\end{table}
 
\noindent The best structure was found at round $39$ and never displaced over the remaining $62$. Just under half the budget produced the answer and the rest produced nothing. None of the $808$ candidates fitted was rejected: the grammar is total, every expression evaluates, so feedback about failed proposals has nothing to work with.
 
\paragraph{What the buffer selected.}
The winning structure carries ten coefficients, of which the run marks three as unidentified:
 
\begin{Verbatim}[fontsize=\footnotesize,breaklines=true,frame=single,framesep=3pt]
c0 + c1*P**c2 + c3*P**c4 + c5*c6**c7/(c8 + t**c9)
              c5 = -3.19  (sd 9.80)
              c6 =  3.69  (sd 8.59)
              c7 =  2.96  (sd 5.50)
\end{Verbatim}
 
\noindent The term \texttt{c5*c6**c7} is a product of three free constants, so it is a constant and the data cannot separate its factors. This is what the identifiability diagnostic is for, and it fires: all three coefficients are named to the critic as undetermined by the data (visible in the evidence of Appendix~\ref{app:prompts}), the critic proposes deleting the term, and the resulting rewrite is fitted as a candidate in its own right. The structure still leads on evidence, with training NMSE $9.5\times10^{-9}$ against $2.4\times10^{-6}$ for the five-coefficient runner-up: the BIC penalty of \eqref{eq:bic} charges for the extra coefficients, and two orders of magnitude of fit outweigh the charge. The trace is a useful illustration of the retention rule doing exactly what Section\ref{sec:critic} says it does -- taking the critic's suggestion as a candidate and nothing more.
 
\paragraph{The final posterior.}
Twenty-four buffer items in $22$ structural families. The top family holds $31\%$ of the weight and the rest is nearly flat, with the next $21$ families between $2.4\%$ and $5.9\%$.
 
\begin{table}[h]
\centering\small
\caption{Top five families of the final buffer, with the whole buffer's weight summing to $1$. Every family is a sum of powers of $P$ and $t$; none contains a saturating term of the kind the ground truth requires.}
\label{tab:buffer}
\begin{tabular}{rlrrr}
\toprule
rank & best expression in family & log-evidence & weight & coeffs \\
\midrule
1 & \texttt{c0+c1*P**c2+c3*P**c4+c5*c6**c7/(c8+t**c9)}     & 26{,}398 & 0.313 & 10 \\
2 & \texttt{c0+c1*P**c2+c3*t**c4+c5*P**c6}                 & 17{,}564 & 0.059 & 7 \\
3 & \texttt{c0+c1*P**c2+c2*P**c3/(c4+t**c5)}               & 17{,}168 & 0.057 & 6 \\
4 & \texttt{c0+c1*P**c2+c3*P**c4+c5*P**c6/(c7+c8*t**c9)}   & 19{,}388 & 0.052 & 10 \\
5 & \texttt{c0+c1*P**c2+c3*P**c4+c5*softplus(c6*t)**c7}    & 18{,}320 & 0.040 & 8 \\
\bottomrule
\end{tabular}
\end{table}
 
\paragraph{Recovery.}
Symbolic accuracy is $0$ at every $k$: $\mathrm{SA}@1$, $@3$, $@5$, $@10$, and $\mathrm{SA}@\text{all}$ over all $22$ families. The credible set holds $19$ families and the true structure is in none of them. The sigmoidal form was never proposed, so this is a property of what the search explored rather than of what it retained -- a distinction the $\mathrm{SA}@k$ metrics of Section\ref{sec:protocol} exist to make.
 
\paragraph{Final intervals.}
The two evaluation splits behave in opposite directions.
 
\begin{table}[h]
\centering\small
\caption{Final metrics, nominal level $0.90$. Widths are absolute; the OOD
interval is roughly $1{,}900$ times wider than the in-distribution one.}
\label{tab:bpg4-metrics}
\begin{tabular}{lrr}
\toprule
 & in-distribution ($\Dtest$) & shifted ($\Dood$) \\
\midrule
NMSE                     & $1.0\times10^{-8}$ & $7.1\times10^{-4}$ \\
$\mathrm{Acc}_{0.1}$     & $0.994$ of points  & $1.000$ of points \\
coverage                 & $0.886$            & $1.000$ \\
coverage gap             & $-0.014$           & $+0.100$ \\
mean interval width      & $1.7\times10^{-4}$ & $3.3\times10^{-1}$ \\
\bottomrule
\end{tabular}
\end{table}
 
This example illustrates why coverage and interval width should be interpreted together. On the OOD test set, the intervals cover all observations, but their mean width is approximately 1,900 times greater than on the ID test set. Thus, perfect OOD coverage comes at the cost of substantially wider intervals. In contrast, ID coverage is 0.886, slightly below the nominal target of 0.90. These results illustrate the importance of evaluating interval width alongside empirical coverage.
 
Note also that the variance decomposition attributes $99.99\%$ of in-distribution predictive variance to parameter uncertainty and only $3\times10^{-8}$ of it to structural disagreement, even though $22$ distinct families are in the buffer. The families agree almost exactly where the data are, and diverge only outside them -- which is the phenomenon the paper is about, visible inside a single run.
 
\paragraph{What the critic did.}
Table~\ref{tab:critic-counters} gives the counters. Over $24$ calls the critic proposed $72$ edits, of which $44$ passed the checks and were fitted; $111$ families to stop proposing, of which $110$ were well formed; and it pointed at evidence that did not exist twice in $255$ claims. It chose to explore on none of the $24$ calls. The proposer went on to propose a banned family $80$ times out of
$701$ proposals made while a ban was active.
 
The last of these is explained by how stagnation is detected. At the call before round $49$ the search had gone nine rounds without improvement, below the threshold of ten, so the critic was not told the search had stalled and chose to refine rather than explore -- on a run whose plateau eventually ran to $62$ rounds. We report this as a limitation in Section\ref{sec:conclusion}.
 
\paragraph{Scope.}
This trace documents the mechanism on one problem: measurements computed from the fits, a critic whose every claim is checked against the evidence it was shown, edits and bans that reach the proposer only after passing those checks, and a retention rule that admits candidates on evidence regardless of where they came from. A single run establishes none of the quantitative claims in this paper; Section\ref{sec:results} reports distributions over problems, seeds and configurations.

\section{Notation}\label{app:notation}
\begin{table}[h]
\centering\small
\caption{Symbols used in Section~\ref{sec:method}, in order of appearance.}
\label{tab:notation}
\begin{tabular}{@{}p{2.6cm}p{10.6cm}@{}}
\toprule
Symbol & Meaning \\
\midrule
$x_i$, $y_i$ & input vector and scalar target of data point $i$ \\
$M$, $M_k$ & a structure: the shape of an equation with its coefficients left free; $M_k$ is candidate $k$ \\
$\theta$, $c_0,c_1,\dots$ & the free coefficients of a structure, and the names they carry in proposals \\
$\hat\theta_k$ & the coefficients of $M_k$ fitted on $\Dfit$ \\
$\Dfit,\Dcid,\Dcext$ & fitting split; labelled calibration split; unlabelled probe region just beyond the data (Table~\ref{tab:splits}) \\
$\Dtest,\Dood$ & the benchmark's in-distribution and shifted test splits, used only at the end \\
$K$ & candidates proposed per round ($8$) \\
$n$ & number of fitting points, $|\Dfit| = 3{,}200$ \\
$d_k$ & number of free coefficients in $M_k$ \\
$\mathrm{RSS}_k$ & residual sum of squares of $M_k$ on $\Dfit$ \\
$\hat\sigma_k^2$ & estimated noise variance of $M_k$, $\mathrm{RSS}_k/n$ with a small floor \\
$\mathrm{BIC}_k$ & Bayesian Information Criterion of $M_k$; lower is better \\
$Z_k$, $\log Z_k$ & evidence (marginal likelihood) of $M_k$ and its logarithm, $\approx -\mathrm{BIC}_k/2$ \\
$\Sigma_k$ & covariance of the fitted coefficients under the Laplace approximation \\
$p(M_k)$, $\lambda$ & prior over structures, $\log p(M_k) = -\lambda d_k$ with $\lambda = 0.05$ \\
$\ell_k$ & score of $M_k$: $\log Z_k + \log p(M_k)$ \\
$B$ & buffer capacity ($24$ structures) \\
$\tau$ & temperature that flattens the weights \\
$w_k$ & tempered weight of $M_k$ in the buffer, used for the spread \\
$\tilde w_k$ & untempered weight ($\tau=1$), used for the predicted mean \\
$B_{\mathrm{eff}}$ & effective number of structures, $(\sum_k w_k^2)^{-1}$; held at $0.35B \approx 8.4$ \\
$\hat y(x)$ & averaged prediction at input $x$ \\
$\sigma_S(x)$ & structural spread: weighted disagreement of the retained structures at $x$ \\
$m$ & size of $\Dcid$ ($800$) \\
$s_i$ & nonconformity score of calibration point $i$ \\
$\alpha$ & tolerated miscoverage ($0.1$, a nominal $90\%$ interval) \\
$\hat q_\alpha$ & conformal quantile of the scores \\
$C(x)$ & reported prediction interval at $x$ \\
\bottomrule
\end{tabular}
\end{table}

\end{document}